\documentclass[lettersize,journal]{IEEEtran}
\usepackage{amsmath,amsfonts}
\usepackage{algorithmic}
\usepackage{algorithm}
\usepackage{array}
\usepackage{bm}
\usepackage[caption=false,font=normalsize,labelfont=sf,textfont=sf]{subfig}
\usepackage{textcomp}
\usepackage{stfloats}
\usepackage{url}
\usepackage{verbatim}
\usepackage{graphicx}
\usepackage{cite}
\usepackage{color}
\usepackage{soul} 
\usepackage{multirow}
\usepackage[numbers,sort&compress]{natbib}

\newcommand{\Rmnum}[1]{\expandafter\@slowromancap\romannumeral #1@}

\usepackage{tikz}
\usepackage{xcolor}
\usepackage{hyperref}
\hypersetup{hypertex=true,
	colorlinks=true,
	linkcolor=black,
	anchorcolor=black,
	citecolor=black}

\definecolor{lime}{HTML}{A6CE39}
\DeclareRobustCommand{\orcidicon}{
	\begin{tikzpicture}
	\draw[lime, fill=lime] (0,0)
	circle[radius=0.16]
	node[white]{{\fontfamily{qag}\selectfont \tiny \.{I}D}};
	\end{tikzpicture}
	\hspace{-2mm}
}
\foreach \x in {A, ..., Z}{%
	\expandafter\xdef\csname orcid\x\endcsname{\noexpand\href{https://orcid.org/\csname orcidauthor\x\endcsname}{\noexpand\orcidicon}}
}

\begin{document}

\title{Metric-aligned Sample Selection and Critical Feature Sampling for Oriented Object Detection}

\author{Peng Sun\hspace{-1.5mm}\orcidA{}, Yongbin Zheng\hspace{-1.5mm}\orcidB{}, Wenqi Wu, Wanying Xu and Shengjian Bai


\thanks{This work was supported in part by the National Natural Science Foundation of China under Grant 62273353. \textit{ (Corresponding author: Yongbin Zheng.)}\par
Peng Sun, Yongbin Zheng, Wenqi Wu ,Wanying Xu and Shengjian Bai are with the College of Intelligence Science and Technology, National University of Defense Technology, 410073 Changsha, China (e-mail: sunpeng@nudt.edu.cn; zybnudt@nudt.edu.cn; wenqiwu\_lit@hotmail.com; wanying\_xu@nudt.edu.cn; baishengjian@nudt.edu.cn).\par
}}

\markboth{IEEE Transactions on Circuits and Systems for Video Technology} 
{Shell \MakeLowercase{\textit{et al.}}: A Sample Article Using IEEEtran.cls for IEEE Journals}


\maketitle

\begin{abstract}

Arbitrary-oriented object detection is a relatively emerging but challenging task. Although remarkable progress has been made, there still remain many unsolved issues due to the large diversity of patterns in orientation, scale, aspect ratio, and visual appearance of objects in aerial images. Most of the existing methods adopt a coarse-grained fixed label assignment strategy and suffer from the inconsistency between the classification score and localization accuracy. First, to align the metric inconsistency between sample selection and regression loss calculation caused by fixed IoU strategy, we represent the regression task as a spatial transformation to evaluate the quality of samples and propose a distance-based label assignment strategy. The proposed metric-aligned selection (MAS) strategy can dynamically select samples according to the shape and rotation characteristic of objects. Second, to further address the inconsistency between classification and localization, we propose a critical feature sampling (CFS) module, which performs localization refinement on the sampling location for classification task to extract critical features accurately. Third, we present a scale-controlled smooth $L_1$ loss (SC-Loss) to adaptively select high quality samples by changing the form of regression loss function based on the statistics of proposals during training. Extensive experiments are conducted on four challenging rotated object detection datasets DOTA, FAIR1M-1.0, HRSC2016, and UCAS-AOD. The results show the state-of-the-art accuracy of the proposed detector.\par

\end{abstract}

\begin{IEEEkeywords}
Arbitrary-oriented object detection, objects with huge diversity, dynamic label assignment, classification feature alignment.\par
\end{IEEEkeywords}

\section{Introduction}
\label{sec_1}

\IEEEPARstart{O}{riented} object detection locates objects of interest with a rotating bounding box and identifies their categories. As a fundamental yet challenging task, oriented object detection has been applied in a wide range of scenarios, such as remote sensing, face recognition, scene text detection and natural scene object detection~\cite{2018DOTA,Raghunandan2019MultiScriptOrientedTD,Han2022AlignDF,Qian2021RSDetPM,yang2021r3det}. Some excellent studies based on angle representation using the oriented bounding box (OBB) with five parameters have emerged recently, such as RoI-transformer~\cite{Ding2019LearningRT}, R$^3$Det~\cite{yang2021r3det} and S$^2$A-Net~\cite{Han2022AlignDF}. Although the above methods have achieved considerable progress, there still suffers heavily from several drawbacks caused by objects with huge diversity, which make it difficult to achieve high-performance detection even with densely set anchors.\par

\begin{figure}[!t]
	\centering
	\subfloat[]{\includegraphics[width=1.6in]{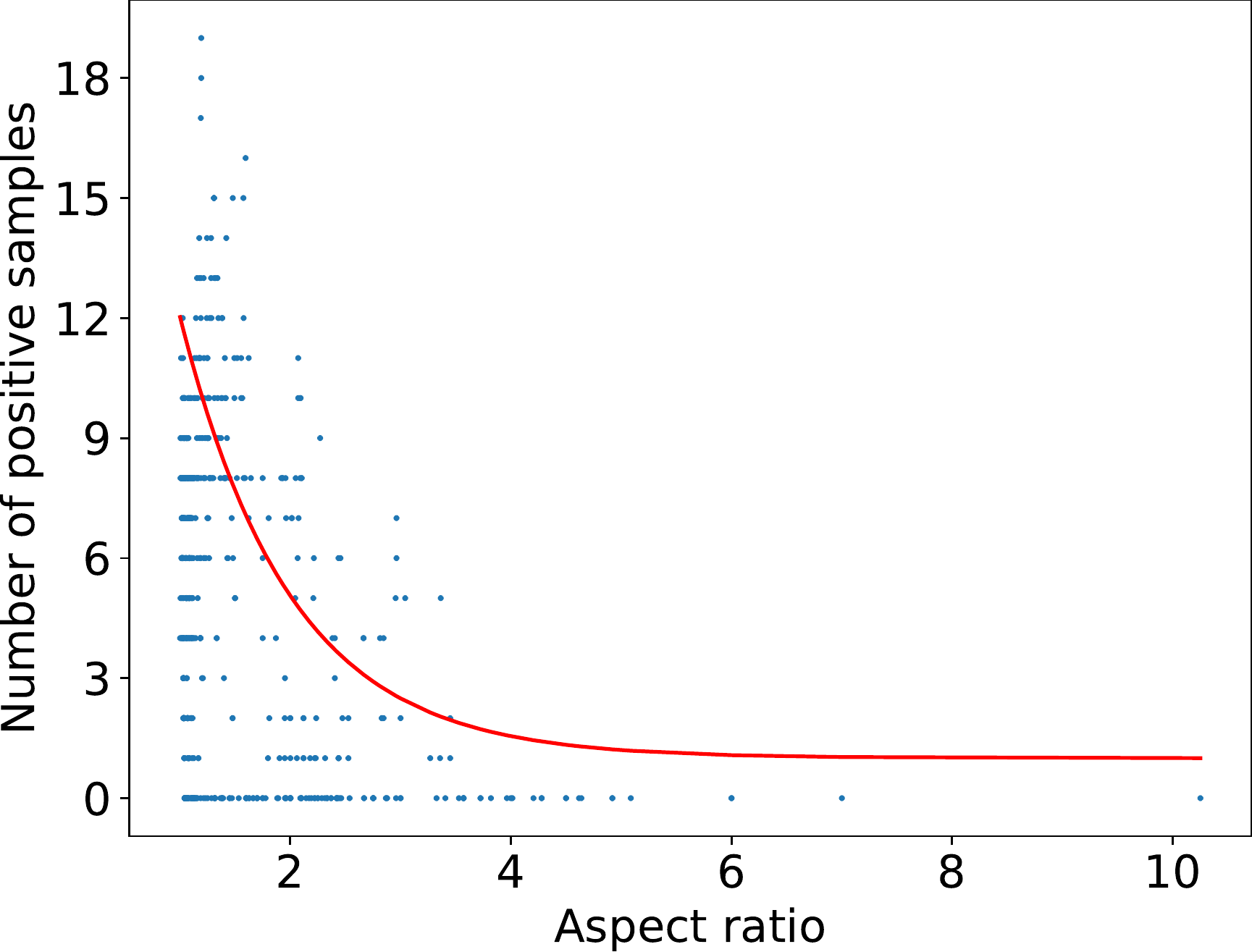}%
		\hspace{0mm}
		\label{(a)}}
	\hfil
	\subfloat[]{\includegraphics[width=1.6in]{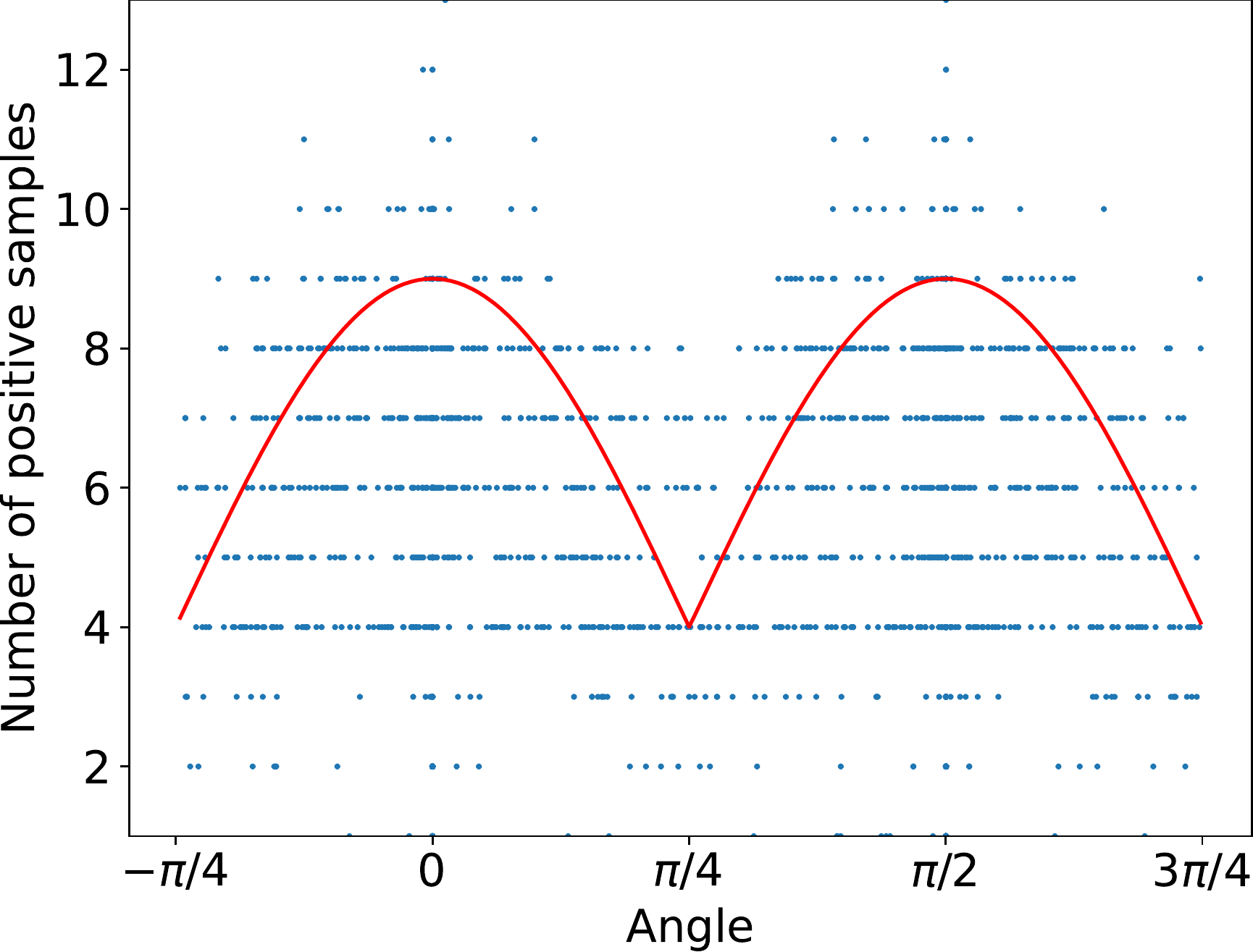}%
		\hspace{0mm}
		\label{(b)}}
	\caption{The relationship between the number of positive samples and (a) aspect ratio and (b) angle of objects in DOTA dataset. It can be seen from the fitting curve that the relationship between the number of positive samples and aspect ratio is monotonic and the relationship between the number of positive samples and angle is periodic.}
	\label{fig_1}
\end{figure}

Firstly, the label assignment of the aforementioned methods selects the positive samples based on IoU, while the regression loss function is calculated according to the four-point distance between the prediction and the ground truth. There exists an inconsistency between the sample selection metric (IoU-based) and regression loss calculation (point distance-based). To address this metric inconsistency, we represent the regression task as a spatial transformation from samples to the ground truth, and analyze the relationship between the sample quality and the shape of objects in sample selection. Based on this, we conduct a distance-based label assignment strategy to select positive samples.\par

Secondly, most of the existing methods adopt a coarse-grained fixed label assignment strategy (MaxIoU), which cannot guarantee sufficient positive samples for a wide variety of hard objects. Fig.~\ref{fig_1}(a) and (b) show the relationship between the number of positive samples and (a) aspect ratio and (b) angle of objects in DOTA dataset. It can be seen that objects with huge aspect ratio variations and arbitrary orientations lack positive samples for regression. To achieve high-performance detection, the label assignment strategy should adopt a dynamic metric for objects with huge variations rather than a fixed one. Many excellent dynamic sample selection strategies have been proposed recently. Zhang et.al~\cite{Zhang2020BridgingTG} noted that the gap in performance between anchor-based and anchor-free methods and constructed a sample selection strategy according to the statistical characteristics of the object. Ming et al.~\cite{ming2021dynamic} adaptively selected high-quality anchors based on their ability to capture critical features. Despite the simplicity and intuitiveness of the dynamic IoU-based metric, objects with large variations were overlooked by existing methods. To solve this issue, some other sample selection methods~\cite{yang2021rethinking,Wang2021ANG,Huang2022AGG} adopt gaussian wasserstein distance (GWD) to measure the difference between samples and the ground truth. For example, Huang et al.~\cite{Huang2022AGG} proposed a general gaussian heatmap labeling to define a positive sample based on two-dimensional (2-D) oriented Gaussian heatmaps. Although GWD-based methods~\cite{Wang2021ANG,Huang2022AGG} reflect the shape and direction characteristics of arbitrary-oriented objects, the form of GWD is complex and cannot measure the distance between square-like objects. To address this problem, we conduct a metric-aligned selection (MAS) strategy, which can dynamically select samples according to the shape and rotation characteristics of objects. Furthermore, we present a scale-controlled smooth $L_1$ loss (SC-Loss), which changes the form of the regression loss function to adaptively fit the distribution change of regression label and select high quality samples for training.\par

\begin{figure}[!t]
	\subfloat[]{\includegraphics[width=1.55in]{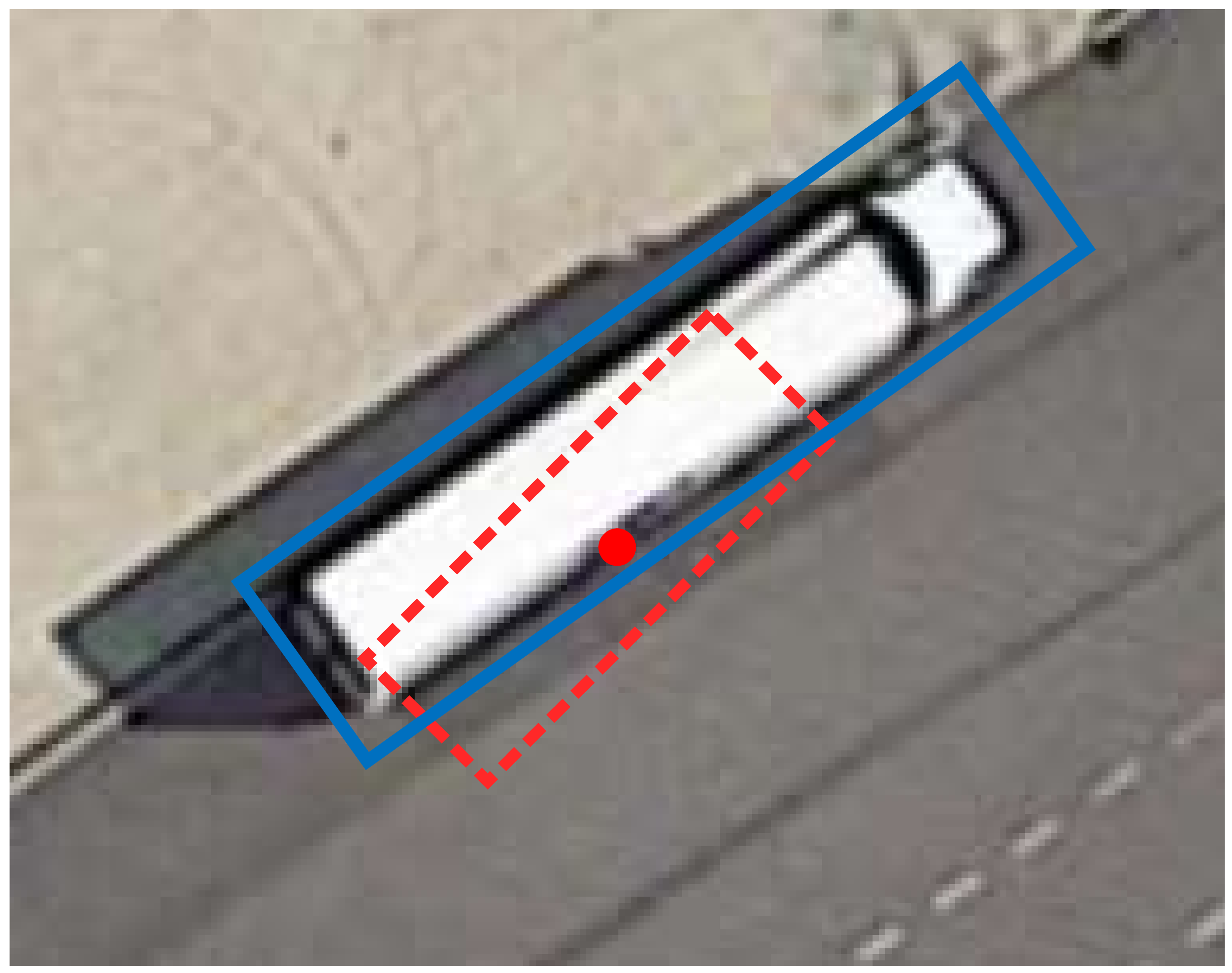}%
		\label{(a)}}
	\hfil
	\subfloat[]{\includegraphics[width=1.55in]{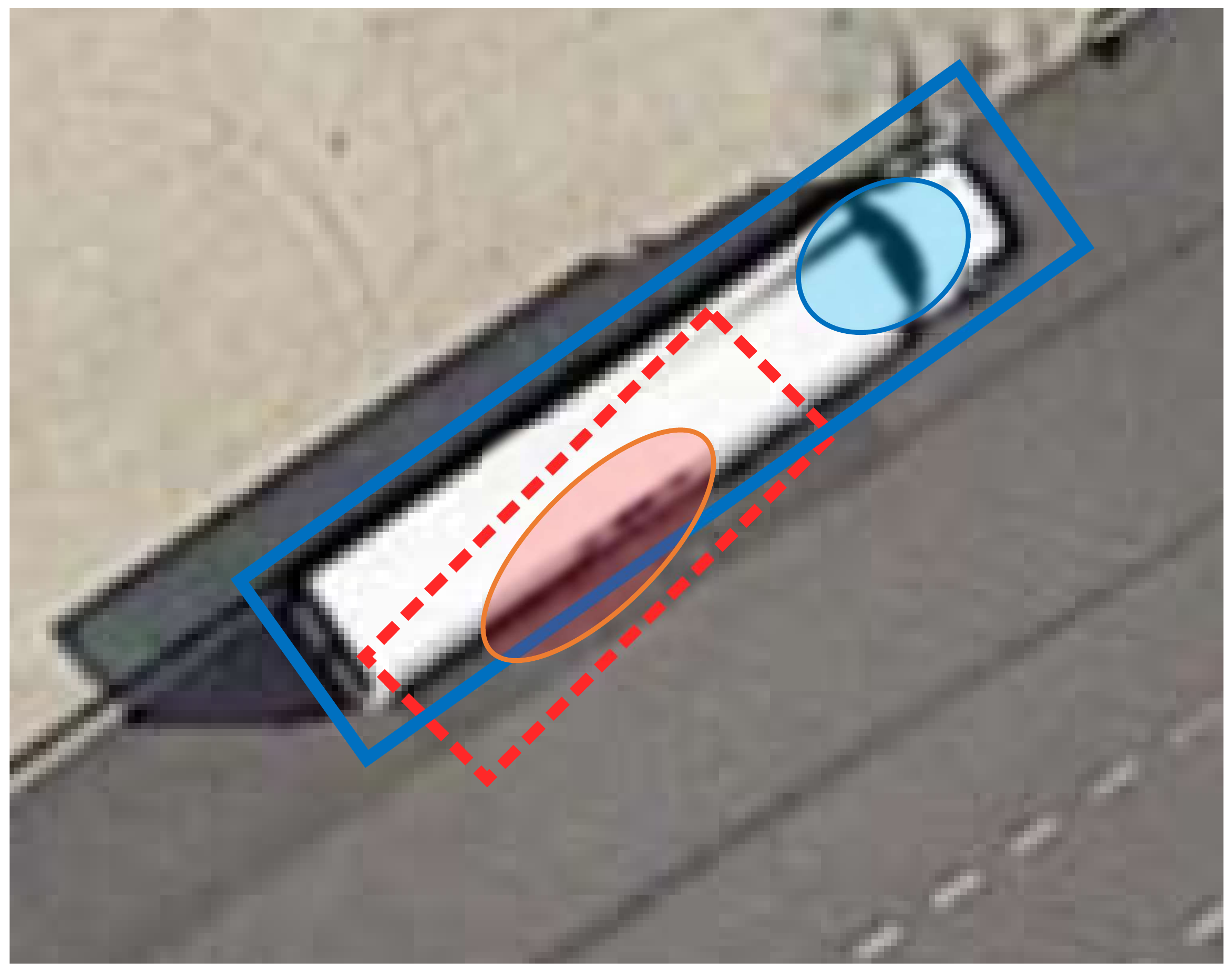}%
		\label{(b)}}
	\caption{The inconsistency between the classification task and the regression task. (a) The refined anchor (red bbox) can achieve accurate location (blue bbox). but (b) the refined anchor (red area) cannot accurately capture the critical features (blue area) for classification.}
	\label{fig_2}
\end{figure}

Thirdly, objects with great visual differences bring a great challenge for classification networks to extract discriminative features. Most object detection methods use the shared features in the classification subnetwork and the regression subnetwork. This leads to inconsistency between the two subnetworks because the classification subnetwork focuses more on rotation invariance features, while the localization subnetwork focuses more on rotation variability features. The great visual differences in oriented object detection exacerbate the inconsistency between the two subnetworks. As shown in Fig.~\ref{fig_2}(a), some methods can refine anchors to accurately locate objects, but not all anchors can capture the critical features used to identify object categories. Only the anchor capturing the critical feature required to identify the object (such as the large vehicle in Fig.~\ref{fig_2}) can achieve the correct classification and the discriminative classification features are included in accurate bboxes. To alleviate the inconsistency between different tasks, we propose a critical feature sampling (CFS) module to perform accurate location to refine the sampling location for classification.\par

In summary, the \textbf{M}etric-aligned \textbf{S}election strategy and \textbf{C}ritical feature \textbf{S}ampling module (\textbf{MSCS}) are proposed for anchor-based oriented object detection. The contributions of this work are summarized as follows:\par

\begin{itemize}
	\item[1)] 
	We propose a distance-based dynamic label assignment strategy MAS to select positive samples for objects with huge diversity. The proposed MAS constructs a distance-based metric to measure the effect of the object's shape and angle characteristic on IoU thresholds.\par
\end{itemize}

\begin{itemize}
	\item[2)]
	We propose a CFS to alleviate the inconsistency between the classification task and the regression task. It can extract critical features for classification task by localization refinement on the sampling location.\par
\end{itemize}

\begin{itemize}
	\item[3)] 
	We propose a SC-Loss to provide high-quality samples for training. The proposed SC-Loss can adaptively select high quality samples by changing the form of regression loss function based on the statistics of proposals during training.\par
\end{itemize}

We report 81.24\%, 46.84\%, 96.47\%, 90.10\% mAPs on the oriented object detection task on the datasets DOTA~\cite{2018DOTA}, FAIR1M-1.0~\cite{sun2022fair1m}, HRSC2016~\cite{liu2017a} and UCAS-AOD~\cite{Zhu2015OrientationRO} respectively, achieving the state-of-the-art in accuracy.\par

The remainder of this paper is organized as follows. The related work is introduced in section~\ref{sec_2}. The details of the proposed methods are introduced in section~\ref{sec_3}. In Section~\ref{sec_4}, extensive experiments are conducted on four challenging remote sensing datasets, including DOTA, FAIR1M-1.0, HRSC2016, and UCAS-AOD. Finally, the conclusion is made in Section~\ref{sec_5}.\par

\begin{figure*}[!t]
	\centering
	\includegraphics[width=6.8in]{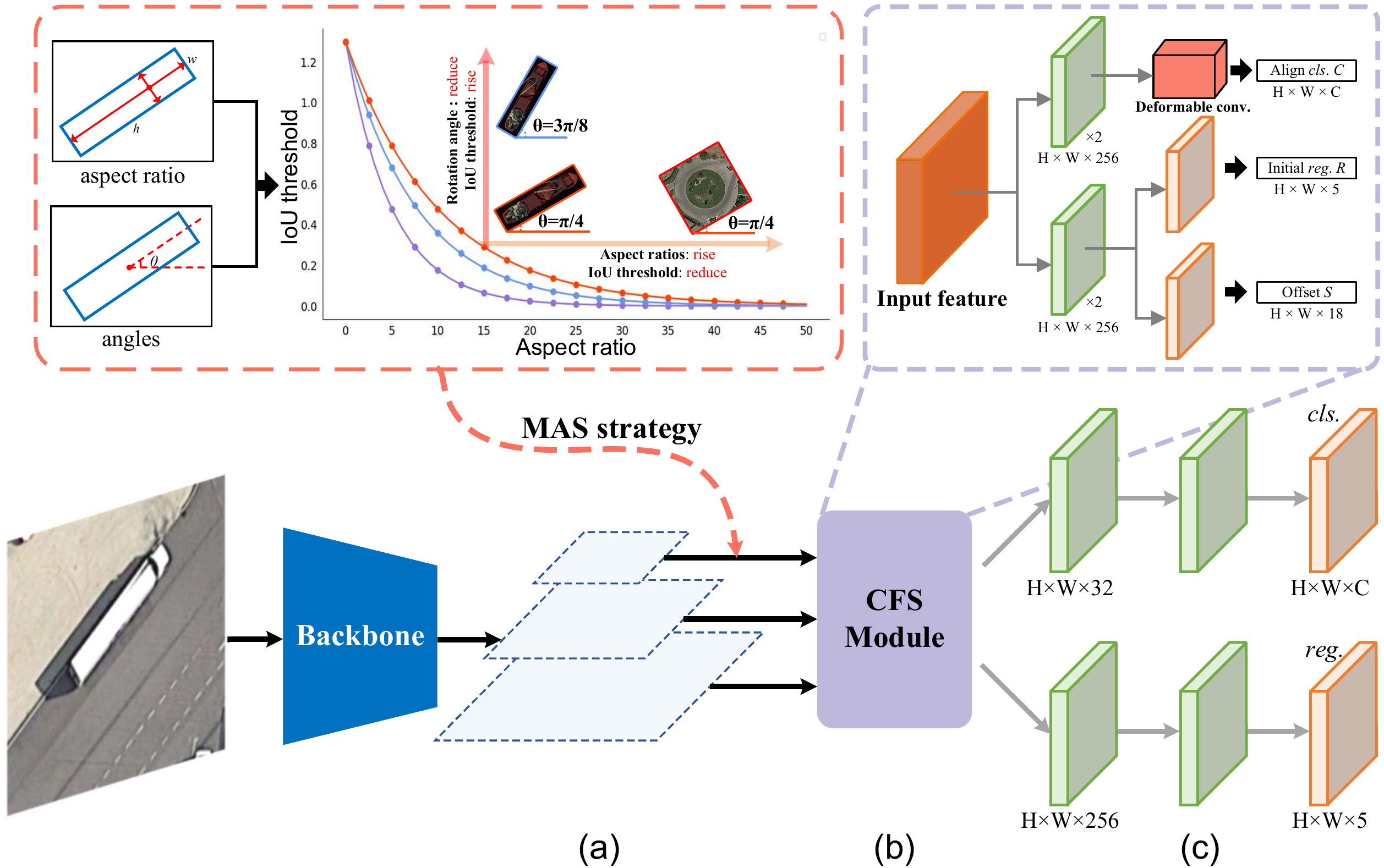}
	\caption{Pipeline of the proposed method. (a) Feature pyramid network. (b) Initial detection head with CFS. (c) Refined detection head. MAS strategy is implemented in the initial detection head to select high-quality samples for objects with different shapes and arbitrary orientations. In MAS, the relationship between the IoU threshold and different aspect ratios and angles can be summarized as: when the aspect ratio is fixed, the rotation angle decreases and the IoU threshold rises (upward arrow); when the rotation angle is fixed, the aspect ratio rises and the IoU threshold decreases (right arrow). The CFS module uses the detection result of the initial detection head to perform localization refinement on the sampling location for \textit{cls}. branch.}
	\label{fig_3}
\end{figure*}

\section{Related Work}
\label{sec_2}

\subsection{Oriented Object Detection in Remote Sensing Images}

The task of oriented object detection adopts the bounding box with angle to extend object detection in remote sensing~\cite{Ma2018ArbitraryOrientedST,Ding2019LearningRT,yang2021r3det,Han2022AlignDF}. With the development of deep neural network-based detectors, current oriented detectors can be classified into two categories: anchor-base oriented detectors and anchor-free oriented detectors. Early anchor-based detector research~\cite{Ma2018ArbitraryOrientedST} introduced the rotated anchor with multiple angles to address arbitrary-oriented objects while involving heavy computations and with no significant improvement. Some methods~\cite{Ding2019LearningRT,yang2021r3det,Han2022AlignDF,Xie2021OrientedRF} retain the horizontal anchor to achieve rotated boxes regression. Ding et al.~\cite{Ding2019LearningRT} proposed the RoI-transformer to learn the transformer from horizontal anchor to oriented anchor. With similar ideas, R$^3$Det~\cite{yang2021r3det} introduced a refinement stage to obtain more accurate features by reencoding the position information. However, they still need a priori settings, such as a multi-scale anchor. S$^2$A-Net~\cite{Han2022AlignDF} adopted a feature alignment module to achieve a higher-quality anchor with few heuristic anchors. Xie et al.~\cite{Xie2021OrientedRF} proposed a new representation of the oriented box and used it in a region proposal network (RPN) to generate high-quality rotating proposals. Although great progress has been achieved in oriented object detection, the detector needs a more flexible sample selection to meet the challenges brought by objects with huge shape variations.\par

Anchor-free methods eliminate heuristic anchors and achieve considerable results. Dynamic refinement network (DRN)~\cite{Pan2020DynamicRN} is based on CenterNet and presents a dynamic refinement network to alleviate the misalignment between axis-aligned neurons and arbitrary-oriented objects. To optimize the regression space, Guo et al.~\cite{Guo2022ConvexHullFA} proposed a convex hull representation method and achieved better performance on DOTA. Many excellent anchor-free methods with point sets have recently emerged. Xu et al.~\cite{Xu2021GlidingVO} designed a four-tuple handcrafted point set to represent objects and achieved point localization. The above methods show that point sets possess great potential for extracting fine-grained representations. However, the point set is manually set on a fixed location, which limits the ability of fine-grained feature extraction. Oriented RepPoints~\cite{li2022oriented} introduced an adaptive point set as the representation of an oriented object. The point set can deal with the semantic and geometric features with more flexibility. Cheng et al.~\cite{Cheng2022AnchorFreeOP} abandoned the operations related to horizontal boxes in the network architecture and generated high-quality object proposals from points.\par

\subsection{Inconsistency Between Regression and Classification}

Most of object detectors~\cite{Ren2015FasterRT,Redmon2018YOLOv3AI,Lin2020FocalLF,Cai2018CascadeRD} adopt two subnetworks to realize classification and location. These subnetworks are usually built by similar structures on shared features, which is called the sibling head. Cheng et al.~\cite{cheng2018revisiting} noted that most of the false detections are caused by the inconsistency between the sibling heads. That is, the inconsistency between regression and classification is that high classification scores may correspond to low-quality locations. To address this issue, Cheng et al.~\cite{cheng2018decoupled} proposed a decoupled classification refinement (DCR), which placed a separate classification network to decouple the classification and regression. Then, the hard false sample was selected, and another classifier was trained to refine the classification results. Other methods alleviate this inconsistency by explicitly aligning the two tasks. RefineDet++~\cite{Zhang2021RefineDetSR} alleviated the mismatch between the spatial features and anchors by adaptively adjusting the spatial features. Wu et al.~\cite{Wu2020RethinkingCA} proposed a double-head method, which uses a fully connected head for classification and a convolution head for location. Feng et al.~\cite{Feng2021TOODTO} proposed a task-aligned one-stage object detection (TOOD) to enhance the interaction between two tasks in a learning-based manner. The inconsistency between classification and regression is more serious in oriented object detection tasks due to the large aspect ratio of objects.\par

\subsection{Label assignment in Object Detection}

Label assignment~\cite{fernandez2018smote,ming2021dynamic,yang2021rethinking,Zhang2020BridgingTG,lv2020iterative,yang2018metaanchor} is to select a number of high quality samples for the detector. Classical one-stage anchor-based detectors~\cite{Redmon2018YOLOv3AI,Lin2017FeaturePN,Lin2020FocalLF} utilize fixed MaxIoU as the matching metric. Recent research~\cite{Zhang2020BridgingTG,Wang2019RegionPB,yang2018metaanchor} proposed that the anchor should be a dynamic metric rather then fixed. Although these dynamic strategies are more efficient than fixed assignment strategies, they have a parameter sensitive problem, such as guided anchoring~\cite{Wang2019RegionPB} and meta anchors~\cite{yang2018metaanchor}. Ming et al.~\cite{ming2021dynamic} proposed a newly defined matching degree that measures the potential of anchors to achieve accurate object detection. Zheng et al.~\cite{Ge2021OTAOT} formulated label assignment as an optimal transport problem. Some sample selection methods define the transformation distance based on the gaussian wasserstein distance. Huang et al.~\cite{Huang2022AGG} proposed a general gaussian heatmap labeling (GGHL) to define a positive sample based on two-dimensional (2-D) oriented Gaussian heatmaps. Wang et al.~\cite{Wang2021ANG} proposed that the IoU is sensitive to tiny objects and then introduced the normalized wasserstein distance (NWD) to address label assignment for tiny objects. However, the weakness of gaussian-based methods~\cite{yang2021rethinking} is that it cannot describes the angle variation between two square objects with similar position. Although existing methods can achieve dynamic sample selection, most of them cannot handle the object with huge variations.\par

\section{The Proposed Method}
\label{sec_3}

We implement the proposed method on the S$^2$A-Net\_$C$, which is one of the baseline structures of S$^2$A-Net. S$^2$A-Net\_$C$ uses normal convolution to replace align convolution, which is easier to verify the performance of the proposed method. The pipeline of the proposed method is shown in Fig.~\ref{fig_3}. It consists of a backbone network, feature pyramid network (FPN), initial detection head with CFS and refined detection head. The initial detection head generates high-quality rotated anchors and extracts aligned features for the refined stage. The refined detection head extracts orientation-invariant features and applies \textit{cls}. branch and \textit{reg}. branch to produce the final detections. The proposed MAS strategy is implemented in the initial detection head to selects high quality samples for objects with different shapes and arbitrary orientations. The proposed CFS module uses the detection result from the initial detection head to perform localization refinement on the sampling location for \textit{cls}. branch to extract critical features accurately. A SC-Loss is designed for the anchor-based bounding box regression, which automatically changes the form of regression loss function based on the statistics of proposals during training.\par

\begin{figure}[!t]
	\centering
	\includegraphics[width=2.8in]{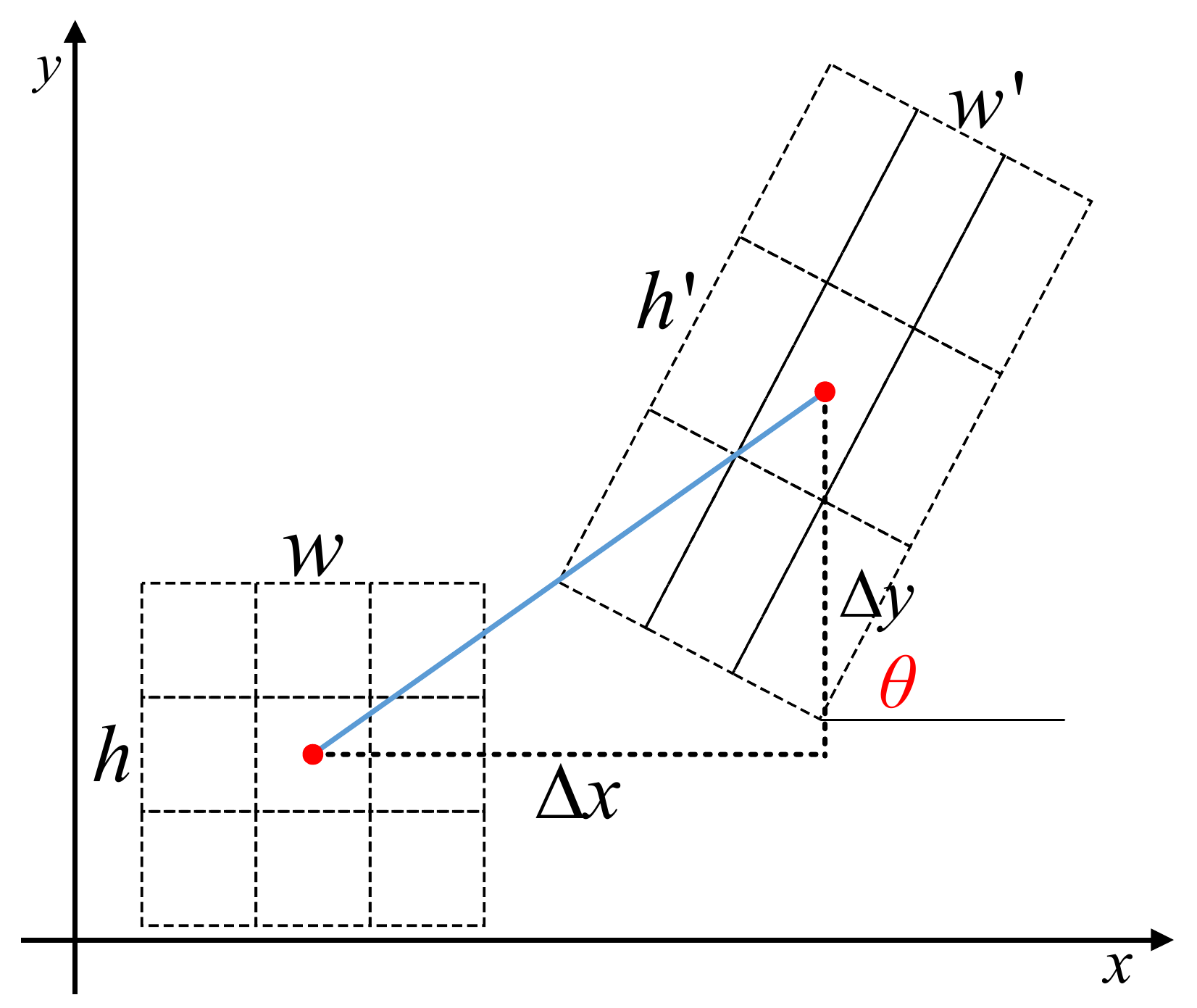}
	\caption{The spatial transformation from the sample to the ground truth.}
	\label{fig_4}
\end{figure}

\subsection{Metric-aligned Selection (MAS) Strategy}

The point distance-based regression loss calculation is a spatial transformation process, but the IoU-based sample selection is not. There exists a metric inconsistency between the regression loss calculation and the sample selection metric. To alleviate the inconsistency, we represent the regression task as a spatial transformation to analyze the relationship between the sample quality and the shape of the objects in sample selection. We summarize that the greater the difference in scale, aspect ratio and angle in spatial transformation, the harder the sample regression, as shown in Fig.~\ref{fig_4}. Due to the well-designed multi-scale anchor set, the impact of the scale variation is negligible in sample selection.\par

In line with the fitting curve in Fig.~\ref{fig_1}(a), the relationship between aspect ratio and sample quality is monotonic, the greater the difference in aspect ratio, the harder it is for the sample to regress. The relationship between angle and sample quality is more complex. In this paper, the samples are all set as square and the angle range is $[-\pi/4, 3\pi/4]$. Due to the geometric symmetry of the sample, when the angle is 0 or $\pi /2$, there is no angle difference between the sample and the ground truth, and the difficulty of sample regression is the smallest; when the angle is $-\pi/4, \pi/4$ or $3\pi/4$, the angle difference is the largest, and the sample is the hardest to regress. The relationship between angle and sample quality is also consistent with the curve in Fig.~\ref{fig_1}(b).\par

Based on the properties of the above analysis, we propose a distance-based dynamic label assignment strategy to allocate sufficient samples for the hard ground truth. In detail, we set a monotonic decreasing function as the weighting factor for the IoU threshold. For a given ground truth $g$, the IoU threshold $IoU^{thres}$ can be defined as:\par

\begin{equation}
\begin{aligned}
IoU_g^{thres} = f({a_g},{\theta _g}) * IoU_g^{init}
\end{aligned}
\label{equ_1}
\end{equation}

\noindent
where $IoU^{init}$ represent the initial IoU threshold. The weighting factor $f(,)$ is mainly related to aspect ratio $a_g$ and angle $\theta_g$, where the larger the aspect ratio, the lower the IoU threshold, which is calculated as follows:\par

\begin{equation}
\begin{aligned}
f({a_g},{\theta _g}) = Co * {e^{ - ({{{a_g}} \mathord{\left/
				{\vphantom {{{a_g}} \gamma }} \right.
				\kern-\nulldelimiterspace} \gamma }) * {\lambda _{{\theta _g}}}}}
\end{aligned}
\label{equ_2}
\end{equation}

\noindent
where $\gamma$ is a weighted parameter, which is defaulted empirically according to the object's aspect ratio. The angle weighted parameter $\lambda$ will be introduced in detail next. Objects with an aspect ratio of 1.5 are the most common objects in remote sensing. The parameter $Co$ is a compensation factor to ensure that these objects have enough samples for regression, which is defined as\par

\begin{equation}
\begin{aligned}
Co = \frac{1}{{{e^{ - ({{1.5} \mathord{\left/
						{\vphantom {{1.5} \gamma }} \right.
						\kern-\nulldelimiterspace} \gamma })}}}}
\end{aligned}
\label{equ_3}
\end{equation}

\noindent
whcih can guarantee that the modulation factor $f(,)$ is 1 when the aspect ratio is 1.5. The angle weighted parameter $\lambda_g$ is defined as:\par

\begin{equation}
\begin{aligned}
{\lambda _g} = \left\{ {\begin{array}{*{20}{c}}
	{ - \frac{1}{2} - {{\sin }^2}\left( { - \theta } \right)}&{ - \frac{\pi }{4} \le \theta  < \frac{\pi }{4}}\\
	{\frac{1}{2} + {{\sin }^2}\left( {\theta  - \frac{\pi }{2}} \right)}&{\frac{\pi }{4} \le \theta  < \frac{{3\pi }}{4},}
	\end{array}} \right.
\end{aligned}
\label{equ_4}
\end{equation}

\noindent
where $0$ and $\pi/2$ are the equilibrium positions of $\lbrack-\pi/4, \pi/4\rbrack$ and $\lbrack\pi/4, 3\pi/4\rbrack$, respectively.\par


Inspired by ATSS~\cite{Zhang2020BridgingTG}, the mean and standard deviation of objects are adopted to set the initial IoU threshold, we have\par

\begin{equation}
\begin{aligned}
IoU_g^{init} = {m_g} + {v_g}
\end{aligned}
\label{equ_5}
\end{equation}

\noindent
where $m_g$ and $v_g$ represent the mean and standard deviation of the IoU between samples and the ground truth, which are defined as:\par

\begin{equation}
\begin{aligned}{m_g} = \frac{1}{N}\sum\limits_{n = 1}^N {{{I}_{i,j}}} ,{\upsilon _g} = \sqrt {\frac{1}{N}\sum\limits_{n = 1}^N {{{({{I}_{i,j}} - {m_g})}^2}} }
\end{aligned}
\label{equ_6}
\end{equation}

\noindent
where $N$ is the number of candidate samples, and ${I}_{i,j}$ is the IoU between the $i$-th ground-truth box and the $j$-th prediction.\par

As shown in the curve in Fig.~\ref{fig_2} is the relationship between IoU threshold and different aspect ratios and angles. It can be seen that the proposed dynamic label assignment strategy is related to the aspect ratio and angle. The relationship between the IoU threshold and different aspect ratios and angles can be summarized as follow: when the aspect ratio is fixed, the farther the rotation angle from the equilibrium position, the smaller the IoU threshold; when the rotation angle is fixed, the larger the aspect ratios, the smaller the IoU threshold.\par

\subsection{Critical Feature Sampling Module}

The task-specific classification features have arbitrary positions and complex deformations, which makes it difficult for standard convolution to accurately capture. Compared with the standard convolution, the deformable convolution~\cite{Dai2017DeformableCN} (DCN) adds an additional offset field to standard convolution to extract features with complex deformations. For each sampling position $\bm{p_0}$, the output feature map of DCN can be expressed as follow:\par

\begin{equation}
\begin{aligned}
\bm{Y}(p_0) = \sum\limits_{r \in \Re ;o \in O} \bm{W}(r) \cdot \bm{X}(p_0 + r + o)
\end{aligned}
\label{equ_7}
\end{equation}

\noindent
where $\Re = \{ ({r_x},{r_y})\} $ is a set of regular grid of standard convolution, $\bm{O}$ is the offset field of DCN. However, DCN optimizes the offset field $\bm{O}$ based on the task-specific, which may result in sampling from the wrong position. In this section, a critical feature sampling (CFS) module is proposed to perform accurate positioning information to optimize the offset field $\bm{O}$.\par

\begin{figure}[!t]
	\centering
	\subfloat[]{\includegraphics[width=3.2in]{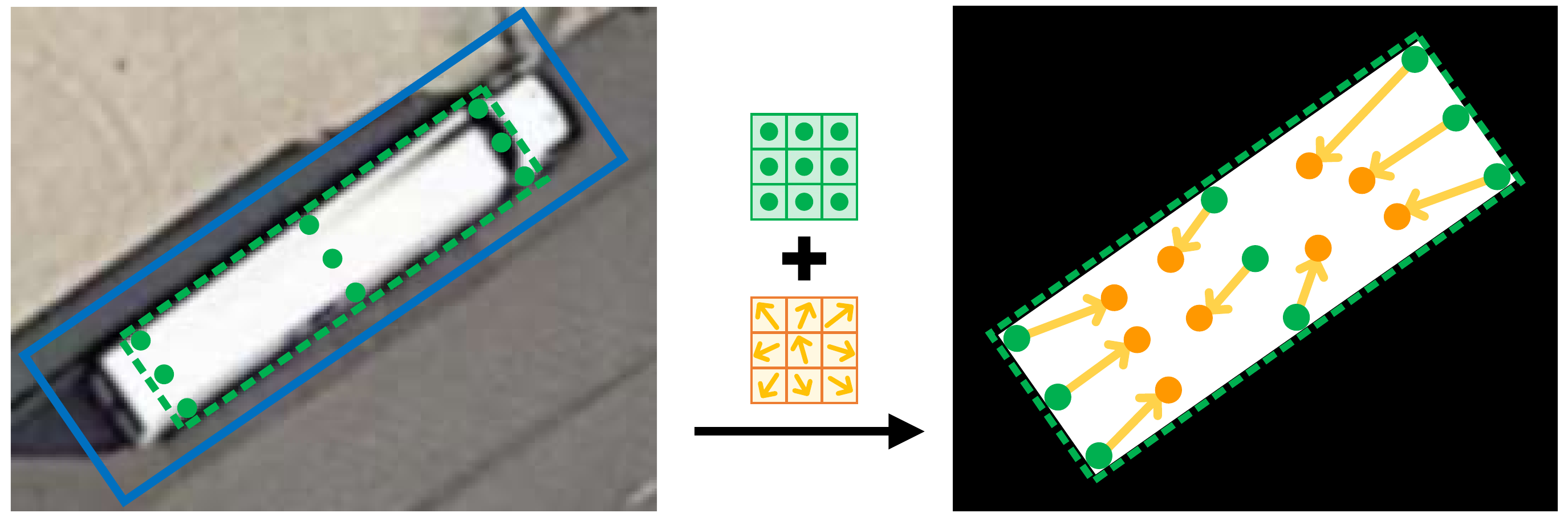}%
		\label{(a)}}
	\hfil	
	\subfloat[]{\includegraphics[width=3.5in]{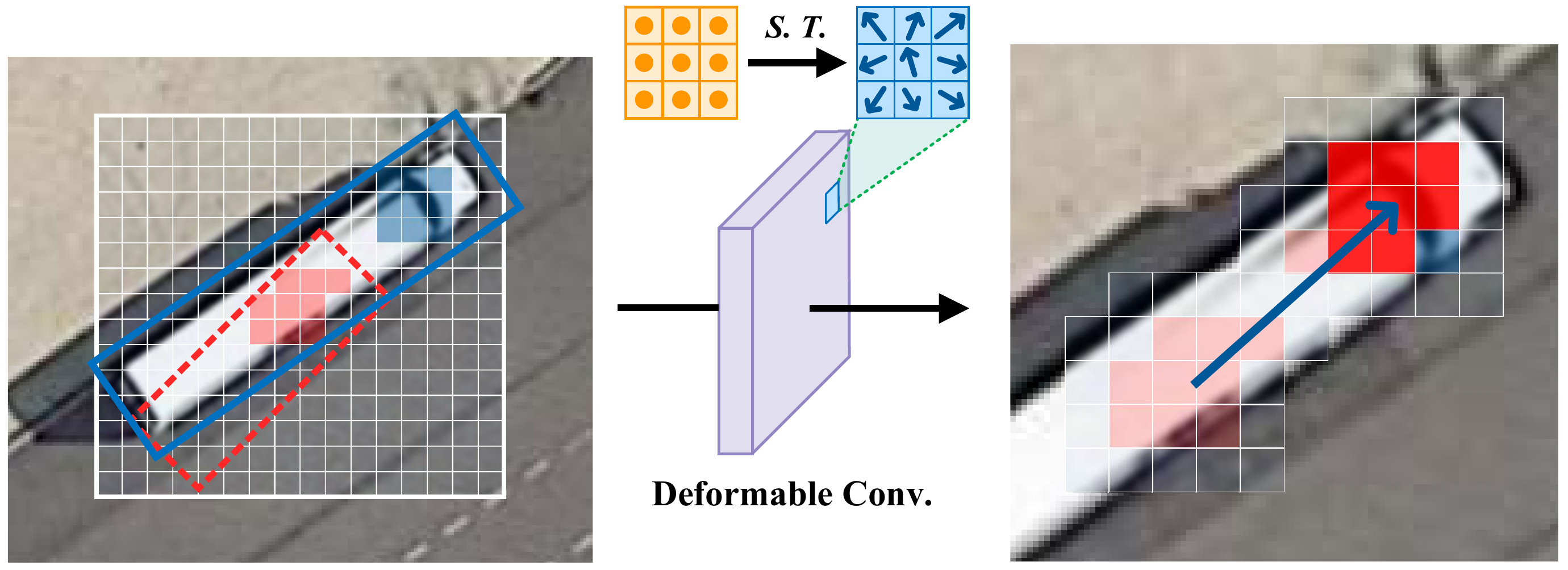}%
		\label{(b)}}
	\hfil	
	\subfloat{\includegraphics[width=3.5in]{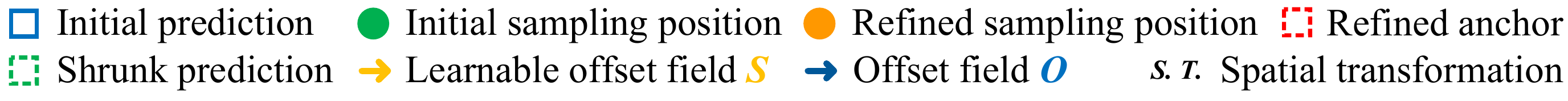}%
	}
	\caption{Overview of the critical feature sampling process: (a) Refined sampling position generation. For the given oriented bounding box (blue), we shrink its boundary by 0.3 to obtain the shrunk box (green) and take the center point, vertex, and midpoint as initial sampling positions (green points). Refined sampling positions (yellow points) can be obtained by initial feature sampling positions and learnable offset field (yellow arrow). (b) Critical feature sampling process. DCN is introduced to achieve the critical feature sampling. And we use spatial transformation to convert the critical feature sampling positions (yellow points) into the sampling position of DCN (blue arrow).}
	\label{fig_5}
\end{figure}

The overview of CFS is shown in Fig.~\ref{fig_5}. As shown in Fig.~\ref{fig_5}(a) is the refined sampling position generation, we first refine the sampling position for classification based on the initial prediction box and learnable offset field $\bm{S}$. Second, a critical feature sampling process is applied to calculate the offset field $\bm{O}$ of each position of DCN based on the refined sampling position (see in Fig.~\ref{fig_5}(b)). Finally, the offset field $\bm{O}$ is fed to DCN to extract discriminative features for classification.\par

\subsubsection{Refined sampling position generation}
For each initial prediction bbox, CFS constructs a set of learnable refined sampling positions to learn the sampling position of critical features, which can be expressed as follow:\par

\begin{equation}
\begin{aligned}
\bm{P} = \left\{ {\bm{p_i}} \right\}_{i = 1}^n,n = 9
\end{aligned}
\label{equ_8}
\end{equation}

\noindent
where \textit{n} is the total number of feature sampling positions. In our work, \textit{n} is set to 9 by default. Refined feature sampling positions can be obtained by initial sampling positions and learnable offset field $\bm{S}$, that is\par

\begin{equation}
\begin{aligned}
\bm{p_i^r} = (x_i^{{\mathop{\rm int}} } + w * \bm{S}_{\Delta {x_i}} ,y_i^{{\mathop{\rm int}} } + h * \bm{S}_{\Delta {y_i}})
\end{aligned}
\label{equ_9}
\end{equation}

\noindent
where $x_{int}$ and $y_{int}$ are the initial sampling positions. The detail of initial sampling positions generation is shown in Fig.~\ref{fig_5}(a). \textit{w} and \textit{h} are the width and height of the bounding box, which are multiplied by the offset to normalize the scale difference. As shown in the network structure of CFS in Fig.~\ref{fig_3}, an additional branch in the regression branch is added to generate the learnable offset field $(\bm{S}_{\Delta {x_i}},\bm{S}_{\Delta {y_i}})$, which can be expressed as follow:\par

\begin{equation}
\begin{aligned}
\left\{ {\begin{array}{*{20}{c}}
	{\bm{R} = \delta \Big({con{v_r}\delta \big({con{v_0}(\bm{F_r})}\big)}\Big)}\\
	{\bm{S} = \delta \Big(con{v_s}\delta \big(con{v_0}(\bm{F_r})\big)\Big)}
	\end{array}} \right.
\end{aligned}
\label{equ_10}
\end{equation}

\noindent
where $con{v_r}$ and $con{v_s}$ are the two branches of two consecutive $con{v_0}$, which are used to obtain the localization features and offset field respectively. $\delta$ represents the activation function. $\bm{R} \in {\mathbb{R}^{{\rm{H}} \times W \times 5}}$ represents the predicted boxes and $\bm{F_r}$ represents the position feature. The channel of $\bm{S}\in{\mathbb{R}^{{\rm{H}} \times W \times 18}}$ represents the offset coordinates of the nine sampling positions.\par

\subsubsection{Critical feature sampling process}

Fig.~\ref{fig_5}(b) is the critical feature sampling process to generate the offset field $\bm{O}$ of DCN. For each position $\bm{p_0}$, the offset field $\bm{O}$ can be calculated by the refined sampling position:\par

\begin{equation}
\begin{aligned}
\bm{O} = {\{ T \cdot p_i^r- p_0 - r\} _{{p_i} \in S,r \in \Re }}
\end{aligned}
\label{equ_11}
\end{equation}

\noindent
where $T$ is a spatial transformation matrix to transform the refined feature sampling position $p_i^r$ into the offset field of DCN. The refined critical features after CFS resampling can be expressed as:\par

\begin{equation}
\begin{aligned}
\bm{F_c^\prime} = con{v_c}({\bm{F_c}} + DCN(\bm{F},\bm{O}))
\end{aligned}
\label{equ_12}
\end{equation}

\noindent
where feature map $\bm{F}$ is the output of the FPN and $\bm{F_c}$ denotes the initial classification feature. $\bm{F_c}^\prime \in {\mathbb{R}^{{\rm{H}} \times W \times C}}$ is the refined feature used for \textit{cls.} branch, which can be obtained by a feature fusion between $\bm{F_c}$ and the output of DCN followed by a convolution.\par

\begin{figure}[!t]
	\centering
	\includegraphics[width=3.0in]{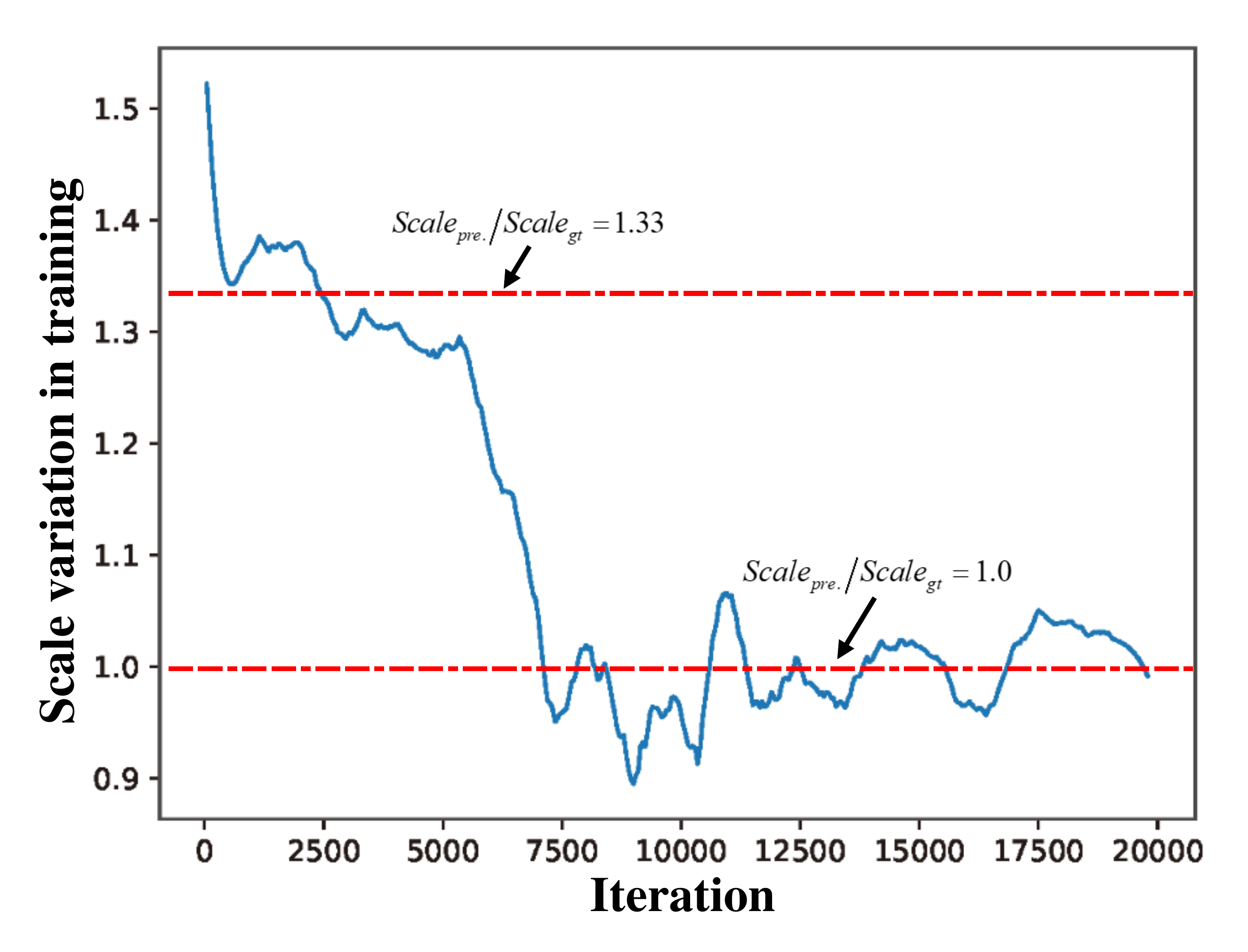}
	\caption{The scale variation between the predicted boxes and the ground truth at different training stages. The quality of the predicted boxes is improved during training.}
	\label{fig_6}
\end{figure}

\subsection{Loss Functions}

In line with the baseline (S$^2$A-Net), long edge definition ($d_{le}$) is adopted to represent an arbitrary-oriented rectangle by five parameters $(x,y,h,w,\theta)$, and the angle $\theta \in [-\pi/4,\ 3\pi/4]$. The multi-task loss of the proposed method is defined as follows.\par

\subsubsection{Total loss}

The total loss is defined as:

\begin{equation}
	\begin{aligned}
		{L_{total}} = {\alpha _{1}}{L_{init.}} + {\alpha _{2}}{L_{ref.}}
	\end{aligned}
	\label{equ_13}
\end{equation}

\noindent
where $L_{init.}$ and $L_{ref.}$ are the multi-task losses of the initial detection head and refined detection head, respectively. The trade-off coefficients $\alpha_{1}$ and $\alpha_{2}$ are set to 1 by default. The multi-task loss is defined as:

\begin{equation}
\begin{aligned}
\begin{array}{l}
L = \frac{{{\lambda _1}}}{N}\sum\limits_{n = 1}^N {t_n^{'}} \sum\limits_{j \in \left\{ {x,y,\omega ,h,\theta } \right\}} {{L_{reg}}\Big(v_{nj}^{'},{v_{nj}}\Big)} \\
+ \frac{{{\lambda _2}}}{N}\sum\limits_{n = 1}^N {{L_{cls}}\Big({p_n},{t_n}\Big)}
\end{array}
\end{aligned}
\label{equ_14}
\end{equation}

\noindent
where hyperparameters $\lambda_1$ and $\lambda_2$ control the trade-off and are set to 1 by default. $N$ indicates the number of anchors, and $t_n^\prime$ is a binary value ($t_n^\prime=1$ for foreground and $t_n^\prime=0$ for background, no regression for background). $v_{nj}^\prime$ represents the predicted offset vectors, and $v_{nj}$ represents the target vector of the ground truth. $t_n$ represents the label of the object, and $p_n$ is the probability distribution of various classes calculated by the sigmoid function. The classification loss $L_{cls}$ and the regression loss $L_{reg}$ are implemented by focal loss~\cite{Lin2020FocalLF} and smooth $L_1$ loss, respectively.\par

\subsubsection{Scale-Controlled smooth $L_1$ Loss}

Fig.~\ref{fig_6} illustrates the scale variation during training, where scale variation is the ratio of the scale of the predicted bbox to the ground truth. It can be seen that the quality of samples is improving with training. Higher quality samples mean better detection results, but the gradient of high-quality samples is smaller under the fixed regression loss function, and the training effect of high-quality samples is reduced. To enhence the effcet of high-quality samples during training, the gradient of regression loss function should change with the improvement of sample quality. The proposed metric in label assignment has been considered the aspect ratio and orientation variation. So, we consider the impact of the scale variation on the smooth $L_1$ loss. Original smooth $L_1$ loss is commonly used to supervise the localization task in object detection, which can be formulated as follows:\par

\begin{equation}
\begin{aligned}
Smoot{h_{L1}}(x) = \left\{ {\begin{array}{*{20}{c}}
	{0.5{{\left| x \right|}^2}/\beta }&{\left| x \right| < \beta}\\
	{\left| x \right| - 0.5\beta }&{otherwise,}
	\end{array}} \right.
\end{aligned}
\label{equ_15}
\end{equation}

\noindent
where $x$ represents the regression error between samples and the ground truth, $\beta$ is a hyperparameter to control the contributions of regression error to the training gradient.\par

Zhang et al.~\cite{zhang2020dynamic} pointed out that different $\beta$ values lead to different curves and gradient of the smooth $L_1$ loss and the smooth $L_1$ loss should change the form to focus on high-quality samples in different training stages. It can be concluded that a smaller $\beta$ means a more saturated gradient, which makes the regression error contribute more to the network training.\par

\begin{figure*}[!t]
	\centering
	\includegraphics[width=7.2in]{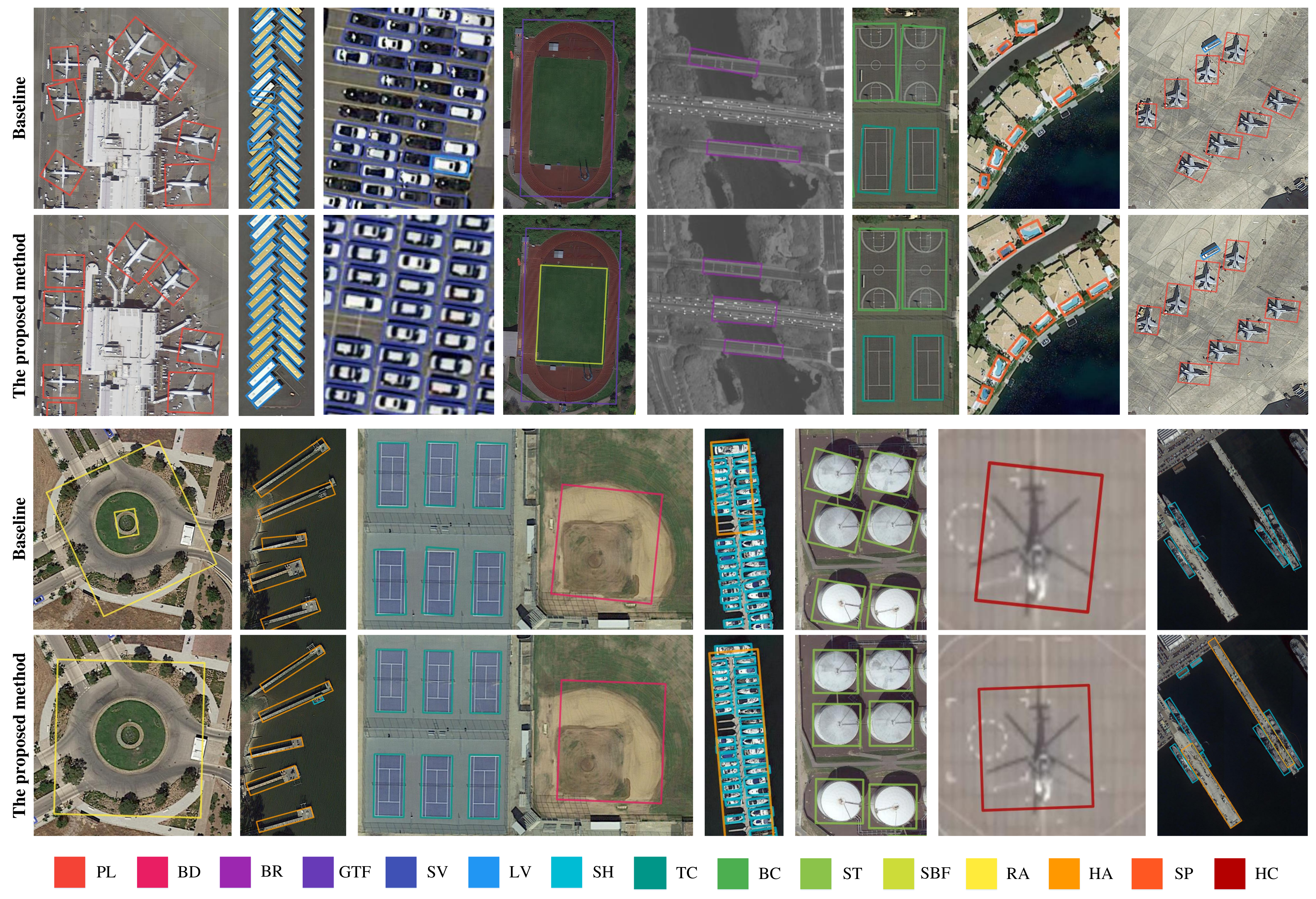}
	\caption{Detection results visualization of baseline with ResNet-101 and the proposed method on \textbf{DOTA}, \textbf{HRSC2016} and \textbf{UCAS-AOD}. Different colored bounding boxes represent instances of different categories (best view in color).}
	\label{fig_7}
\end{figure*}

In this paper, a scale-controlled smooth $L_1$ loss (SC-Loss) is proposed, which consider the scale variation in training process and sets a dynamic hyperparameter $\beta_{scale}$ for Smooth $L_1$ loss. During network training, SC-Loss will change $\beta_{scale}$ according to adaptively fit the distribution change of regression label. For every iteration, we first calculate the scale similarity between proposals and the ground truth and select the $K_{th}$ smallest value. Due to more outliers in regression labels, we calculate the median of the normalized scale instead of the average and use it to update the $\beta_{scale}$ every iteration. In this way, SC-Loss pays more attention to samples with large-scale variation and obtains a high-quality regressor.\par

\section{Experiments and Discussions}
\label{sec_4}

We perform experiments on four well-known public aerial image benchmarks, that is, DOTA, FAIR1M-1.0, HRSC2016, and UCAS-AOD. All experiments are implemented by mmdetection\footnote{\href{https://github.com/open-mmlab/mmdetection}{https://github.com/open-mmlab/mmdetection}}\cite{chen2019mmdetection} on a server with NVIDIA GeForce RTX 3090 with 24G memory. In this subsection, we present the benchmark dataset, network and parameter implementation, ablation study and experimental results.\par

\subsection{Dataset}

\textbf{DOTA} is a challenging dataset for oriented object detection in remote sensing images. It contains 2,806 satellite optical images with sizes ranging from 800$\times$800 to 4000$\times$4000 pixels, with a total of 188,282 instances. Instances in DOTA exhibit a wide variety of scales, orientations, and shapes. These images are then annotated by experts using 15 common object categories. The object categories include plane (PL), baseball diamond (BD), bridge (BR), ground track field (GTF), small vehicle (SV), large vehicle (LV), ship (SH), tennis court (TC), basketball court (BC), storage tank (ST), soccer-ball field (SBF), roundabout (RA), harbor (HA), swimming pool (SP), and helicopter (HC). DOTA uses random selections of the original image as follows: half of the original images are randomly selected as the training set, 1/6 as the validation set, and 1/3 as the testing set. It is noted that the image size in DOTA is too large, and we divide the training and validation into 1024$\times$1024 with a stride of 200. An official website\footnote{\href{https://captain-whu.github.io/DOTA/dataset.html}{https://captain-whu.github.io/DOTA/dataset.html}} is provided for submitting the results and performance verification.\par

\textbf{FAIR1M-v1.0} is a recently published large remote sensing dataset. It consists of 15,266 high-resolution images with a resolution of 300 to 800 from different platforms, and more than 1 million instances for fine-grained object recognition. All instances in the FAIR1M-1.0 dataset are annotated with 5 categories and 37 sub-categories by oriented bounding boxes.\par

\textbf{HRSC2016} is a high-resolution remote sensing dataset for ships with oriented bounding box annotations. The dataset contains 1061 images in total, which are collected from six well-known harbors. The image sizes range from 300$\times$300 to 1500$\times$900. These images are divided into 436 for the training set, 181 for the validation set and 444 for the testing set. All images were resized to 800$\times$512 for training and testing.\par

\textbf{UCAS-AOD} is an aerial aircraft and car detection dataset with 1,510 images collected from Google Earth, in which 1,000 images contain planes (with 7,482 instances) and 510 images contain cars (with 7,114 instances). We randomly divide it into a training set and a test set at a ratio of 7:3.\par

\begin{table*}[!t]\scriptsize
	\renewcommand\arraystretch{1.1}
	\centering
	\caption{Ablation study of each structure in our proposed methods on the \textbf{DOTA} dataset and the \textbf{HRSC2016} dataset.}
	\setlength{\tabcolsep}{1.5mm}{
		\begin{tabular}{c|ccc|c|c|c|c|c|c|c|c|c|c|c|c|c|c|c|c}\hline
			\textbf{Dataset}&\textbf{MAS}&\textbf{CFS}&\textbf{SC-Loss}&\textbf{mAP}&\textbf{SH}&\textbf{PL}&\textbf{BD}&\textbf{BR}&\textbf{GTF}&\textbf{SV}&\textbf{LV}&\textbf{TC}&\textbf{BC}&\textbf{ST}&\textbf{SBF}&\textbf{RA}&\textbf{HA}&\textbf{SP}&\textbf{HC}\\\hline
			
			\multirow{5}{*}{DOTA}			
			&-&-&-&71.89&78.90&89.35&80.40&48.68&69.37&77.36&72.47&90.79&79.97&84.11&61.50&62.57&64.17&69.15&49.61\\
			&$\surd$&&&73.06&79.38&89.41&77.66&\textbf{52.99}&67.60&\textbf{77.99}&75.03&90.83&84.25&84.80&61.34&62.89&66.92&\textbf{69.32}&\textbf{55.40}\\
			&&$\surd$&&73.09&79.42&89.29&81.60&51.41&69.68&77.70&73.34&90.83&83.75&85.39&\textbf{62.13}&65.23&65.32&69.65&51.64\\
			&&&$\surd$&72.31&\textbf{87.69}&76.23&49.46&52.91&76.56&74.55&\textbf{79.52}&90.85&85.36&85.63&60.85&61.67&66.74&69.03&50.57\\
			&$\surd$&$\surd$&$\surd$&\textbf{74.43}&86.62&\textbf{89.43}&\textbf{82.15}&52.18&\textbf{71.51}&76.66&78.51&\textbf{90.86}&\textbf{85.70}&\textbf{85.55}&60.20&\textbf{66.26}&67.23&69.03&54.50\\\hline
			
			\multirow{5}{*}{HRSC2016}			
			&-&-&-&-&85.60\\
			&$\surd$&&&-&89.20\\
			&&$\surd$&&-&90.01\\
			&&&$\surd$&-&87.30\\
			&$\surd$&$\surd$&$\surd$&-&\textbf{90.41}\\\hline
			
		\end{tabular}
	}
	\label{tab_1}
\end{table*}

\subsection{Implementation Details}

Experiments are conducted by ResNet101 with FPN as the basic network by default unless otherwise specified. For each level of pyramid features, only one squared anchor is placed in the corresponding location. All models are trained by the SGD optimizer using 4 GPUs with a total of 8 images per mini batch. The hyperparameters of the initial learning rate, momentum and weight decay are set to 0.01, 0.9 and 0.0001, respectively. The framework is trained by 12, 12, 36 and 120 epochs on DOTA, FAIR1M-1.0, HRSC2016 and UCAS-AOD, respectively.\par

\subsection{Ablation Study}

To verify the effectiveness of the proposed method, we conduct a series of comparison experiments on DOTA and HRSC2016. Several baselines are adopted to study the impact of the individual structure proposed in this paper. The results are illustrated in Table~\ref{tab_1}. The performance (AP for each category and overall mAP) demonstrates that each proposed structure achieved considerable improvements in detection performance on different datasets.\par

\begin{table}[!t]\scriptsize
	\renewcommand\arraystretch{1.1}
	\centering
	\caption{Analysis of different hyperparameter $\gamma$ of MAS on the \textbf{DOTA} dataset. 6-mAP means the performance of the categories listed.}
	\setlength{\tabcolsep}{2.2mm}{
		\begin{tabular}{c|cccccc|cc}\hline
			\textbf{$\gamma$}&\textbf{BR}&\textbf{GTF}&\textbf{LV}&\textbf{SH}&\textbf{BC}&\textbf{HA}&\textbf{6-mAP}&\textbf{mAP}\\\hline
			
			3&51.54&64.45&74.03&78.91&83.96&65.97&69.81&70.06\\			
			4&52.80&66.90&73.86&79.04&80.68&65.80&69.85&71.94\\			
			5&\textbf{52.99}&67.60&\textbf{75.03}&\textbf{79.38}&\textbf{84.25}&\textbf{66.92}&\textbf{71.03}&\textbf{73.06}\\			
			6&50.14&\textbf{68.84}&73.78&79.02&82.63&65.91&70.05&71.95\\			
			7&46.30&64.66&71.60&78.71&80.21&64.49&67.66&72.01\\			
			\hline			
		\end{tabular}
	}
	\label{tab_2}
\end{table}

\subsubsection{Hyper-parameters}

To find suitable hyperparameter settings of the MAS, parameter sensitivity experiments are conducted, and the result is shown in Table~\ref{tab_2}. The AP of each category listed and their mAP can obtain the best when $\gamma$ is set to 5 on the DOTA. Considering the distribution differences between the above datasets, the weighted parameter $\gamma$ for the MAS is also set to 5 for FAIR1M-1.0 and UCAS-AOD due to the similar distribution with DOTA, and HRSC2016 is set to 6 due to the more exaggerated aspect ratio objects.\par

\subsubsection{Effect of the MAS}

As shown in Table~\ref{tab_1}, S$^2$A-Net\_$C$ with the MAS achieves 73.06\% mAP on the DOTA dataset and 89.20\% AP on HRSC2016, increasing by 1.17\% and 3.20\%, respectively. Table~\ref{tab_1} shows that performances achieved 79.38\%, 52.99\%, 75.03\%, 66.92\%, 67.60\% and 84.25\% for some specific categories, such as SH, BR, LV, HA, GTF and BC. The results of the above categories demonstrate the effectiveness of the MAS for objects with huge diversity.\par

The result in Table~\ref{tab_3} is the comparison between the MAS and other advanced label assignments based on S$^2$A-Net\_$C$. Compared with other dynamic label assignment methods, such as ATSS and SA-S, the proposed label assignment improves the mAP to 73.06\% on DOTA. In particular, experiments on MAS with $\lambda = 1$ are conducted here. When $\lambda = 1$, the MAS achieves 72.22\% mAP on DOTA, which is also better than other methods and slightly lower than MAS. Especially for objects with large aspect ratios and scales, such as BR, LV, and BC, the accuracy was significantly improved by 52.99\%, 75.03\%, and 84.25\%, respectively and the 3-mAP achieved best for 70.76\%. The results show the importance of the angle modulation factor.\par

\begin{table}[!t]\scriptsize
	\renewcommand\arraystretch{1.1}
	\centering
	\caption{Results of different label assignment for objects with large aspect ratio and scale on \textbf{DOTA} dataset. 3-mAP means the performance of the categories listed.}
	\setlength{\tabcolsep}{2.4mm}{
		\begin{tabular}{c|ccc|cc}\hline
			\textbf{Sample Selection Strategy}&\textbf{BR}&\textbf{LV}&\textbf{BC}&\textbf{3-mAP}&\textbf{mAP}\\\hline
			
			MaxIoU&48.29&71.83&79.91&66.68&71.57\\
			ATSS~\cite{Zhang2020BridgingTG}&50.12&72.75&81.2&68.04&71.71\\
			SA-S~\cite{hou2022shape}&50.30&72.43&84.11&68.95&71.90\\
			MAS with $\lambda = 1$~(Ours)&50.75&74.07&83.46&69.43&72.22\\
			MAS~(Ours)&\textbf{52.99}&\textbf{75.03}&\textbf{84.25}&\textbf{70.76}&\textbf{73.06}\\\hline			
		\end{tabular}
	}
	\label{tab_3}
\end{table}

\begin{table}[!t]\scriptsize
	\renewcommand\arraystretch{1.1}
	\centering
	\caption{Experiments of different baseline and different anchor set on \textbf{HRSC2016}.}
	\setlength{\tabcolsep}{2.8mm}{
		\begin{tabular}{c|ccc|c}\hline
			\textbf{Baseline}&\textbf{MaxIoU}&\textbf{MAS}&\textbf{Anchor}&\textbf{AP}\\\hline
			
			\multirow{3}{*}{Rotated RetinaNet}
			&$\surd$&&\textbf{9}&82.67\\
			&&$\surd$&\textbf{9}&77.60\\
			&&$\surd$&\textbf{1}&88.59\\\hline
			
			\multirow{2}{*}{S$^{2}$A-Net\_$C$}
			&$\surd$&&\textbf{1}&85.60\\
			&&$\surd$&\textbf{1}&89.20\\\hline			
		\end{tabular}
	}
	\label{tab_4}
\end{table}

\subsubsection{Effect of the number of anchors in the MAS}

As is known, setting the anchor is important for object detection. To further explore the impact of the number of anchors for the MAS, we perform different baseline (rotated RetinaNet and S$^{2}$A-Net\_$C$) on HRSC2016. As shown in Table~\ref{tab_4}, rotated RetinaNet adopted MaxIoU with nine anchors per location and achieved a mAP of 82.67\%. Surprisingly, when the MAS was adopted for the baseline, the accuracy drops by 5.07\%. In deeper experiments, we set one anchor per location for the MAS, and the accuracy increased significantly and achieved 88.59\%. Further analysis shows that too much prior information (number of anchors) is redundant for the MAS. Experiments also show that a well-designed label assignment can achieve competitive performance with few anchors.\par

\begin{table*}[!t]\scriptsize
	\renewcommand\arraystretch{1.3}
	\centering
	\caption{Comparison with state-of-the-art methods on the OBB-based task of the \textbf{DOTA}. MS mean Multi-Scale training and testing.}
	\setlength{\tabcolsep}{1.2mm}{
		\begin{tabular}{c|c|c|c|c|c|c|c|c|c|c|c|c|c|c|c|c|c|c}\hline
			
			\textbf{Methods}&\textbf{Backbone}&\textbf{MS}&\textbf{PL}&\textbf{BD}&\textbf{BR}&\textbf{GTF}&\textbf{SV}&\textbf{LV}&\textbf{SH}&\textbf{TC}&\textbf{BC}&\textbf{ST}&\textbf{SBF}&\textbf{RA}&\textbf{HA}&\textbf{SP}&\textbf{HC}&\textbf{mAP}\\\hline
			
			\textbf{Anchor-free:}\\\hline
			
			O2-DNET~\cite{Dai2022AO2DETRAO}&H-104&$\surd$&89.31&82.14&47.33&61.21&71.32&74.03&78.62&90.76&82.23&81.36&60.93&60.17&58.21&66.98&61.03&71.04\\
			
			P-RSDet~\cite{Zhou2020ArbitraryOrientedOD}&R-101&$\surd$&88.58&77.83&50.44&69.29&71.1&75.79&78.66&90.88&80.1&81.71&57.92&63.03&66.3&69.77&63.13&72.30\\
			
			BBAVECTORS~\cite{Yi2021OrientedOD}&R-101&$\surd$&88.63&84.06&52.13&69.56&78.26&80.40&88.06&90.87&87.23&86.39&56.11&65.62&67.10&72.08&63.96&75.36\\
			
			POLARDET~\cite{Zhao2020PolarDetAF}&R-101&$\surd$&89.65&\textbf{\textcolor{red}{87.07}}&48.14&70.97&78.53&80.34&87.45&90.76&85.63&86.87&61.64&\textbf{\textcolor{blue}{70.32}}&71.92&73.09&67.15&76.64\\
			
			AOPG~\cite{Cheng2022AnchorFreeOP}&R-50&$\surd$&89.88&85.57&60.90&81.51&78.70&85.29&88.85&90.89&87.60&87.65&71.66&68.69&\textbf{\textcolor{red}{82.31}}&77.32&73.10&80.66\\
			
			\hline
			
			\textbf{One-stage:}\\\hline
			
			
			
			R$^3$Det~\cite{yang2021r3det}&R-101&&88.76&83.09&50.91&67.27&76.23&80.39&86.72&90.78&84.68&83.24&61.98&61.35&66.91&70.63&53.94&73.79\\
			
			R$^4$Det~\cite{sun2020r4}&R-152&&88.96&85.42&52.91&73.84&74.86&81.52&80.29&90.79&86.95&85.25&64.05&60.93&69.00&70.55&67.76&75.84\\
			
			DAL~\cite{ming2021dynamic}&R-50&$\surd$&89.69&83.11&55.03&71.00&78.3&81.9&88.46&90.89&84.97&87.46&64.41&65.65&76.86&72.09&64.35&76.95\\	
			
			GWD~\cite{yang2021rethinking}&R-152&$\surd$&89.06&84.32&55.33&77.53&76.95&70.28&83.95&89.75&84.51&86.06&73.47&67.77&72.60&75.76&74.17&77.43\\
			
			GGHL~\cite{Huang2022AGG}&D-53&$\surd$&\textbf{\textcolor{blue}{89.74}}&85.63&44.5&77.48&76.72&80.45&86.16&90.83&\textbf{\textcolor{blue}{88.18}}&86.25&67.07&69.4&73.38&68.45&70.14&76.95\\
			
			S$^2$A-NET~\cite{Han2022AlignDF}&R-101&&88.70&81.41&54.28&69.75&78.04&80.54&88.04&90.69&84.75&86.22&65.03&65.81&76.16&73.37&58.86&76.11\\
			
			ADT-Det~\cite{zheng2021adt}&R-152&$\surd$&89.61&84.59&53.18&80.05&78.31&80.86&88.22&90.82&84.80&86.89&69.97&66.78&76.18&72.10&60.03&77.43\\
			
			S$^2$A-NET&R-101&$\surd$&89.28&84.11&56.95&79.21&\textbf{\textcolor{blue}{80.18}}&82.93&89.21&90.86&84.66&87.61&71.66&68.23&78.58&78.20&65.55&79.15\\
			
			\hline
			
			\textbf{Two-stage:}\\\hline
			
			
			ROI-TRANS.~\cite{Ding2019LearningRT}&R-101&$\surd$&88.64&78.54&43.44&75.92&68.81&73.68&83.59&90.74&77.27&81.46&58.39&53.54&62.83&58.93&47.67&69.56\\
			
			
			Gliding Vertex~\cite{Xu2021GlidingVO}&R-101&&89.64&85.00&52.26&77.34&73.01&73.14&86.82&90.74&79.02&86.81&59.55&70.91&72.94&70.86&57.32&75.02\\
			
			
			CENTERMAP~\cite{Wang2021LearningCP}&R-101&$\surd$&89.83&84.41&54.60&70.25&77.66&78.32&87.19&90.66&84.89&85.27&56.46&69.23&74.13&71.56&66.06&76.03\\
			
			FPN-CSL~\cite{yang2020arbitrary}&R-152&$\surd$&\textbf{\textcolor{red}{90.25}}&85.53&54.64&75.31&70.44&73.51&77.62&90.84&86.15&86.69&69.60&68.04&73.83&71.10&68.93&76.17\\
			
			SCRDet++~\cite{Qian2021RSDetPM}&R-101&$\surd$&90.05&84.39&55.44&73.99&77.54&71.11&86.05&90.67&87.32&87.08&69.62&68.90&73.74&71.29&65.08&76.81\\
			
			REDET~\cite{Han2021ReDetAR}&RER-50&$\surd$&88.81&82.48&60.83&\textbf{\textcolor{blue}{80.82}}&78.34&\textbf{\textcolor{red}{86.06}}&88.31&90.87&\textbf{\textcolor{red}{88.77}}&87.03&68.65&66.9&79.26&\textbf{\textcolor{red}{79.71}}&\textbf{\textcolor{blue}{74.67}}&80.10\\
			
			Orinented R-CNN~\cite{Xie2021OrientedRF}&R-50&$\surd$&89.84&85.43&\textbf{\textcolor{blue}{61.09}}&79.82&79.71&85.35&88.82&90.88&86.68&\textbf{\textcolor{red}{87.73}}&\textbf{\textcolor{blue}{72.21}}&\textbf{\textcolor{red}{70.80}}&82.42&78.18&74.11&\textbf{\textcolor{blue}{80.87}}\\
			
			\hline
			
			
			S$^2$A-Net+MSCS~(Ours)&R-101&&89.72&82.57&55.36&73.21&76.99&81.05&87.61&90.85&85.82&85.84&65.79&65.44&76.84&72.73&58.86&76.58\\
			
			S$^2$A-Net+MSCS~(Ours)&R-101&$\surd$&89.80&\textbf{\textcolor{blue}{86.23}}&58.13&80.12&79.34&84.97&\textbf{\textcolor{blue}{88.67}}&90.86&85.82&87.14&65.36&69.12&\textbf{\textcolor{blue}{79.05}}&\textbf{\textcolor{blue}{78.99}}&69.31&79.53\\
			
			S$^2$A-Net+MSCS~(Ours)&SWIN-TINY&$\surd$&89.69&85.80&\textbf{\textcolor{red}{61.76}}&\textbf{\textcolor{red}{82.33}}&\textbf{\textcolor{red}{80.59}}&\textbf{\textcolor{blue}{85.97}}&\textbf{\textcolor{red}{88.80}}&\textbf{\textcolor{blue}{90.89}}&86.98&\textbf{\textcolor{blue}{87.68}}&\textbf{\textcolor{red}{74.37}}&69.67&78.51&78.49&\textbf{\textcolor{red}{77.08}}&\textbf{\textcolor{red}{81.24}}\\			
			
			\hline			
		\end{tabular}
	}
	\label{tab_6}
\end{table*}




\begin{figure}[!t]
	\centering
	\includegraphics[width=3.5in]{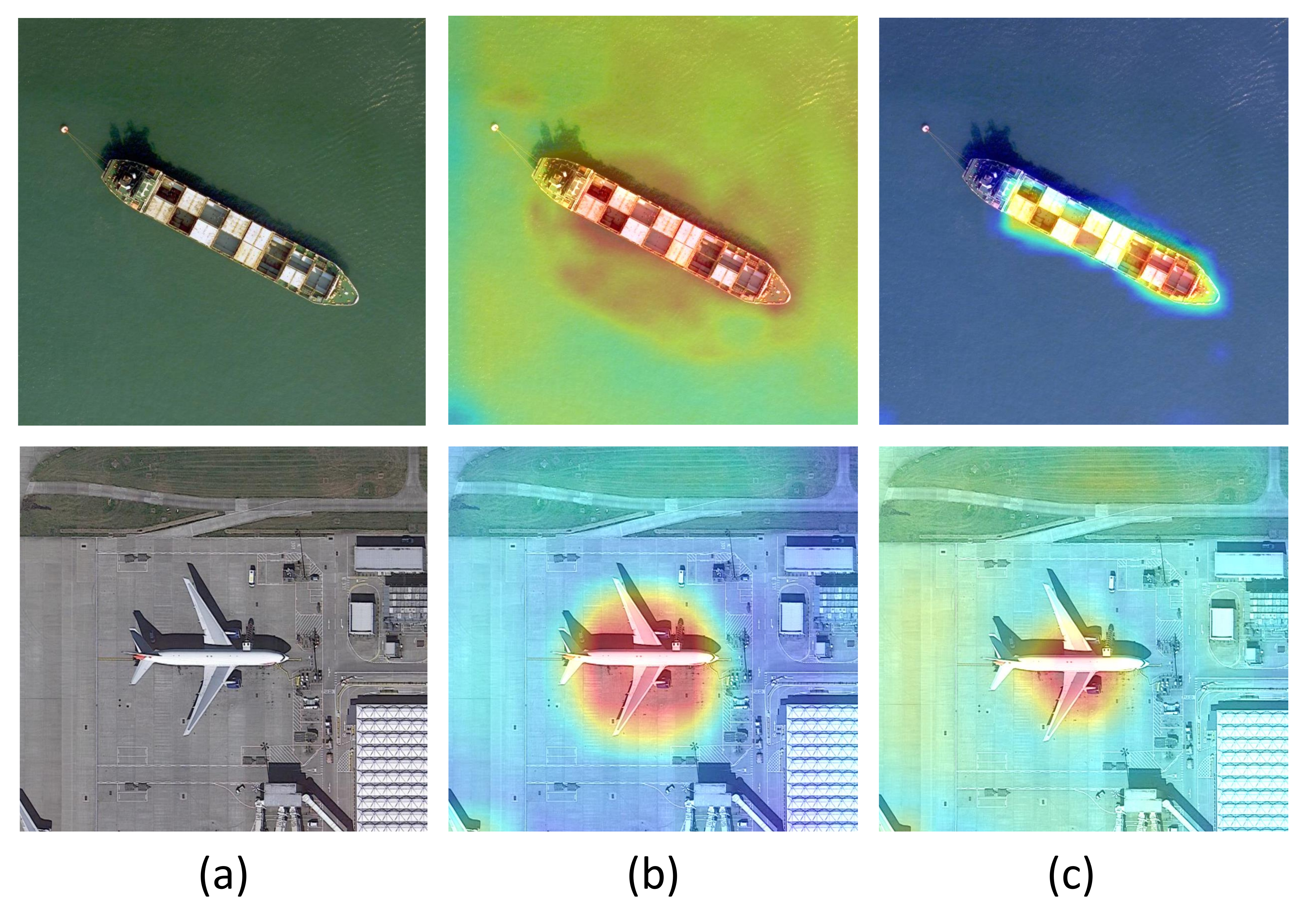}
	\caption{Visualization results of the critical feature for classification. (a) Images. (b) Shared feature. (c) Classification feature generated by CFS. Compared with the shared feature, the classification feature refined by CFS can better focus on the critical features of their respective tasks.}
	\label{fig_8}
\end{figure}

\subsubsection{Effect of the CFS method}

Table~\ref{tab_1} shows the ablation study results for each strategy on DOTA and HRSC2016. All models selected S$^2$A-Net\_$C$ as the baseline. The mAP performance achieved 73.09\% and 90.01\% with CFS implemented and improved by 1.20\% and 4.41\% on DOTA and HRSC2016. This result proved that the accurate location can refine the detector to extract more accurate classification features and improve the detector performance. Some visualization results are shown in Fig.~\ref{fig_8}. It can be seen that discriminative features required for classification are often concentrated in the local part of objects, such as at the middle of planes or at the stern and bow of ships. The heatmap generated by CFS can accurately capture the area of \textit{cls}. sensitive critical features.\par

\subsubsection{Effect of SC-Loss}

As shown in Table~\ref{tab_1}, we evaluate the effectiveness of scale-controlled smooth $L_1$ loss. Categories such as SH, BR, LV, and HA on DOTA have the characteristics of large aspect ratios or large-scale differences. As can be seen, the proposed SC-Loss achieved 72.31\% mAP on DOTA and the APs of the above categories improved by 8.79\%, 4.23\%, 7.05\%, and 2.57\%, respectively. Experiments were also performed on HRSC2016, which achieved 1.70\% AP improvement, to convincingly prove the effectiveness of the SC-Loss. These experimental results demonstrated that SC-Loss can improve the detector's scale adaptability.\par

\subsection{Comparisons with State-of-the-art Detectors}

\subsubsection{Results on DOTA}

We compared the proposed approach with the state-of-the-art methods on DOTA-v1.0 OBB Task, the results of various methods are shown in Table~\ref{tab_6}. It is worth noting that the results are from submitting the test results to the official website. As shown in Table~\ref{tab_6}, S$^2$A-Net and R$^3$Det are advanced rotation detectors that achieve state-of-the-art performance. The proposed method based on S$^2$A-Net with R-101-FPN further improve the performance by 0.47\% and 2.83\% without any data augmentation. In addition, with multi-scale training and testing strategies, our method achieved the best result of 79.53\% under R-101-FPN and 81.24\% under Swin-Tiny~\cite{Liu2021SwinTH}, which outperforms the advanced two-stage detector Oriented R-CNN~\cite{Xie2021OrientedRF} and anchor-free detector AOPG~\cite{Cheng2022AnchorFreeOP}. Facing the challenging DOTA dataset, the proposed S$^2$A-Net-based method achieved the best results in 6 categories (the results are marked red in Table~\ref{tab_6}): BR, GTF, SV, SH, SBF, HC. In particular, we achieved the above results by utilizing only one anchor (no-redundant scale and aspect ratio) at each feature point. The qualitative detection results of the baseline method (S$^2$A-Net) and our proposed method are visualized in Fig.~\ref{fig_7}. Compared with baseline, the proposed method produces more accurate location and less false predictions when detecting on objects with different shape and arbitrary orientation.\par

\begin{table*}[!t]\scriptsize
	\renewcommand\arraystretch{1.2}
	\centering
	\caption{Comparison with state-of-the-art methods on the \textbf{FAIR1M-v1.0} dataset.}
	\setlength{\tabcolsep}{1.5mm}{
		\begin{tabular}{c|ccccccc}\hline
			\textbf{Methods}&Gliding Vertex~\cite{Xu2021GlidingVO}&RetinaNet~\cite{Lin2020FocalLF}&Cascade R-CNN~\cite{Cai2018CascadeRD}&Faster R-CNN~\cite{Ren2015FasterRT}&RoI Trans.~\cite{Ding2019LearningRT}&Orinented R-CNN~\cite{Xie2021OrientedRF}&S$^2$A-Net+MSCS~(Ours)\\
			\hline
			\textbf{mAP}(\%)&29.92&30.67&31.18&32.12&35.29&45.60&\textbf{46.84}\\
			\hline		
		\end{tabular}
	}
	\label{tab_7}
\end{table*}

\begin{figure*}[!t]
	\centering
	\includegraphics[width=7.0in]{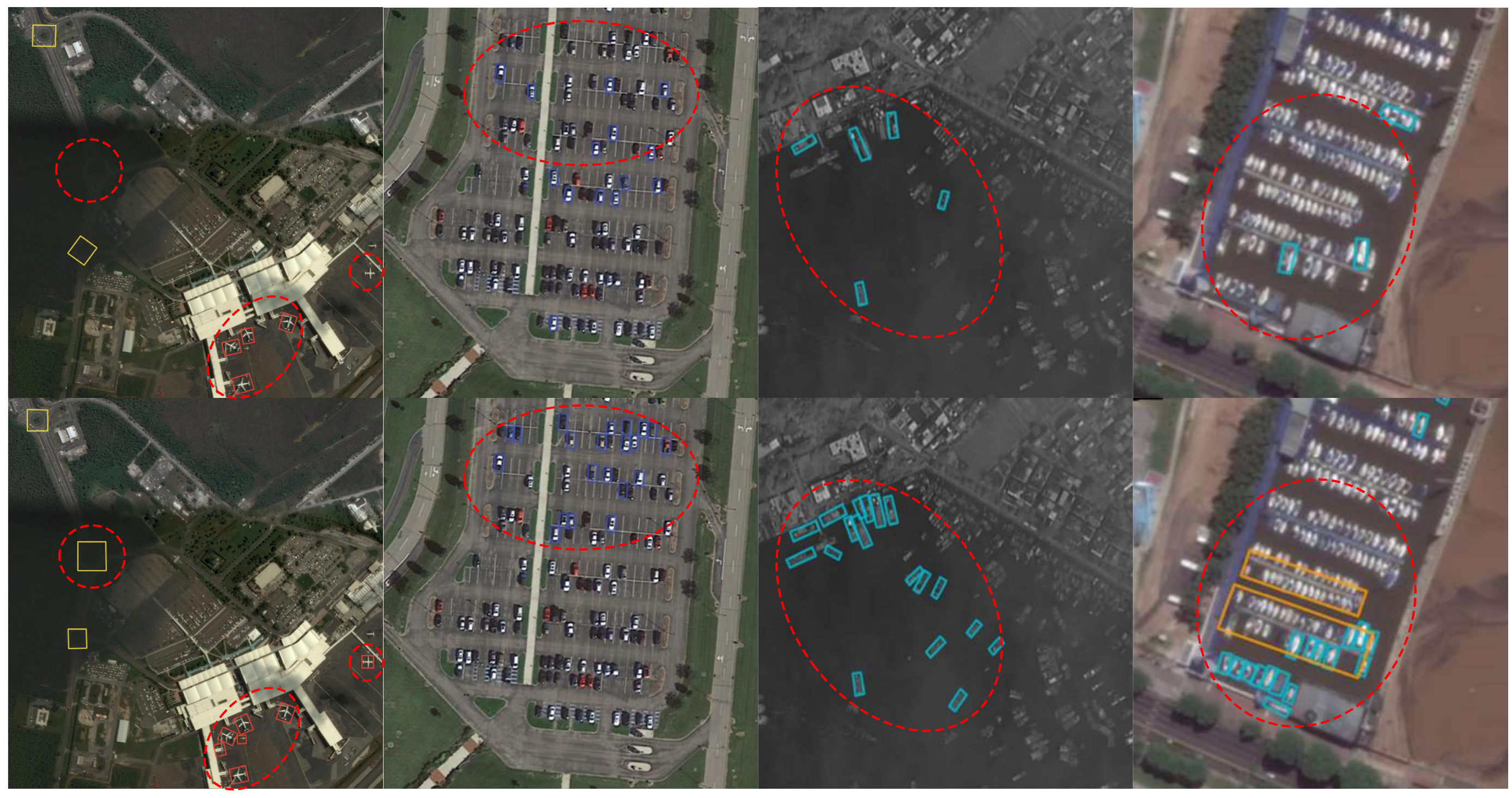}
	\caption{Detection results visual comparisons on the \textbf{FAIR1M-1.0} dataset. For each pair of detection results, the top one is baseline, and the bottom one is the proposed method. The area surrounded by the red dotted line is the focus of attention.}
	\label{fig_9}
\end{figure*}

\subsubsection{Results on FAIR1M-1.0}

FAIR1M-1.0 is a challenging multi-categories dataset. We compare the proposed method against 6 other models on the FAIR1M-v1.0 dataset. Experimental results in Tab.\ref{tab_7} show that the proposed method achieves 46.84\% mAP, which is the best performance among the compared detectors. As shown in Fig.~\ref{fig_9}, we also perform some direct visual comparisons and the area of the red dotted line in figure is the main concern area. Objects in remote sensing image are always surrounded by complex environments, such as blurred picture quality, complex backgrounds, and cloud and fog occlusion, which lead to false detection or missed detection. Although the proposed method has a lower rate of missed detection and false detection, it still has a large improvement space. Improving detectors to deal with low-quality image object detection is the focus of our next research.\par

\subsubsection{Results on HRSC2016}

The HRSC2016 dataset contains many slender ship objects with arbitrary orientations, which poses a considerable challenge to detection. It is noted that the performance under PASCAL VOC2012 metrics was higher than that under PASCAL VOC2007 metrics. To be fair, our experiment on HRSC2016 uses two metrics. We compare the proposed methods with other state-of-the-art methods in Table~\ref{tab_8}. The experimental results show that the proposed method achieved 89.32\% accuracy with RetinaNet as the baseline, which improves the detection performance by 6.43\%. The proposed method based on S$^2$A-Net was superior to the other methods and achieved 90.48\% under AP50 and 96.90\% under AP75. The performance of existing methods on HRSC2016 has reached a relatively high level, the proposed method only improved by 1.89\% on the baseline. The experimental results further illustrate the advantages of the proposed techniques in object detection with a large aspect ratio.\par

\begin{table}[t]\scriptsize
	\renewcommand\arraystretch{1.1}
	\centering
	\caption{Comparisons with state-of-the-art Methods on \textbf{HRSC2016} dataset. The AP50 mean evaluated the result under PASCAL VOC2007 metrics, and the AP75 mean evaluated the result under PASCAL VOC2012 metrics.}
	\setlength{\tabcolsep}{4mm}{
		\begin{tabular}{c|c|cc}\hline
			
			\textbf{Methods}&\textbf{Backbone}&\textbf{AP50}&\textbf{AP75}\\\hline
			RRD~\cite{Liao2018RotationSensitiveRF}&VGG16&84.3&-\\
			RepPoints~\cite{Yang2019RepPointsPS}&R-101&85.16&-\\
			ROI-TRANS.~\cite{Ding2019LearningRT}&R-101&86.2&-\\
			DRN~\cite{Pan2020DynamicRN}&H-104&-&92.7\\
			GLIDING VECTORS~\cite{Xu2021GlidingVO}&R-101&88.2&-\\
			BBAVECTORS~\cite{Yi2021OrientedOD}&R-101&88.44&-\\
			R$^3$Det~\cite{yang2021r3det}&R-101&89.26&96.01\\
			DCL-R$^3$Det~\cite{Yang2021DenseLE}&R-101&89.46&96.41\\
			RetinaNet-H&R-101&82.89&89.27\\
			S$^2$A-NET~\cite{Han2022AlignDF}&R-101&90.17&95.01\\\hline			
			RetinaNet+MSCS~(Ours)&R-101&89.32&93.16\\
			S$^2$A-Net+MSCS~(Ours)&R-101&\textbf{90.48}&\textbf{96.90}\\\hline			
		\end{tabular}
	}
	\label{tab_8}
\end{table}

\subsubsection{Results on UCAS-AOD}

The car and plane in UCAS-AOD are usually small and distributed in any direction and surrounded by complex scenes. Further experiments are conducted in UCAS-AOD, and the results are shown in Table~\ref{tab_9}. The proposed rotating object detector achieved competitive performances of 90.10\% for AP50 and 52.16\% for AP75. Note that our method works best under both AP50 and AP75, which reveals that the MAS with SC-Loss helps the detector distinguish high-quality samples and achieve high-quality object detection.\par

\begin{table}[!t]\scriptsize
	\renewcommand\arraystretch{1.2}
	\centering
	\caption{Comparisons with the performance of different methods on \textbf{UCAS-AOD} dataset.}
	\setlength{\tabcolsep}{3.5mm}{
		\begin{tabular}{c|cc|cc}\hline
			
			\textbf{Methods}&\textbf{Car}&\textbf{Airplane}&\textbf{AP50}&\textbf{AP75}\\\hline			
			YOLOv3~\cite{Redmon2018YOLOv3AI}&74.63&89.52&82.08&-\\			
			RepPoints~\cite{Yang2019RepPointsPS}&83.02&89.34&86.18&-\\			
			RetinaNet~\cite{Lin2020FocalLF}&84.64&90.51&87.57&-\\			
			Faster-RCNN~\cite{Ren2015FasterRT}&86.87&89.86&88.36&47.08\\			
			ROI-TRANS.~\cite{Ding2019LearningRT}&88.02&90.02&89.02&50.54\\
			RIDet-Q~\cite{Ming2022OptimizationFA}&88.5&89.96&89.23&-\\
			RIDet-O~\cite{Ming2022OptimizationFA}&88.88&90.35&89.62&-\\
			DAL~\cite{ming2021dynamic}&89.25&90.49&89.87&-\\
			SASM~\cite{hou2022shape}&89.56&90.42&90.00&-\\\hline
			RetinaNet+MSCS~(Ours)&88.52&\textbf{90.58}&89.55&50.02\\
			S$^2$A-Net+MSCS~(Ours)&\textbf{89.68}&90.52&\textbf{90.10}&\textbf{52.16}\\\hline			
		\end{tabular}
	}
	\label{tab_9}
\end{table}


			
			
			
			
			
			
			
			
			
			

\subsubsection{Speed Comparison}

Comparative experiments for speed and accuracy are performed on the DOTA dataset. All experiments are based on R-50-FPN and the size of input images is 1024$\times$1024. The hardware platform of testing is a single NVIDIA GeForce RTX 3090 GPU with batch size of 1. The detailed experimental results of different methods are shown in Table~\ref{tab_10}. The single-stage S$^2$A-Net can achieve 74.12\% mAP and 15.3 FPS. We implement our S$^2$A-Net-based detector by mmdetection and spent 1.4 hours (12 epochs) to optimize the model. The proposed method can achieve an 75.35\% mAP and a 14.2 FPS. It can be seen that the proposed detector improved on the S$^2$A-Net obtains better performance with comparable speed.\par

\section{Conclusion}
\label{sec_5}

In this paper, we first propose a MAS strategy to achieve metric-alignment between sample selection and regression loss calculation. It can dynamically select high-quality samples according to the shape and rotation characteristic of objects. Second, to further address the inconsistency between classification and localization, we propose a CFS module, which performs localization refinement on the sampling location for classification task to extract accurate critical features. Third, we present a SC-Loss to adaptively select high quality samples by changing the form of regression loss function based on the statistics of proposals during training. Extensive experiments are conducted on four challenging rotated object detection datasets. The results show the state-of-the-art accuracy of the proposed detector.\par

In label assignment, we only used spatial transformation to analyze the relationship between the sample quality and the shape of the object, which is a qualitative analysis. In the future, we will study how to model the distance of spatial transformation to quantitatively evaluate the difficulty of sample regression.\par

\begin{table}[!t]\scriptsize
	\renewcommand\arraystretch{1.2}
	\centering
	\caption{Speed versus accuracy on the \textbf{DOTA} dataset.}
	\setlength{\tabcolsep}{4.5mm}{
		\begin{tabular}{c|c|c|c}\hline			
			\textbf{Methods}&\textbf{Framwork}&\textbf{FPS}&\textbf{mAP}\\\hline
			RetinaNet-O~\cite{Lin2020FocalLF}&One-stage&\textbf{16.1}&68.43\\
			S$^2$A-Net~\cite{Han2022AlignDF}&One-stage&15.3&74.12\\
			Faster R-CNN-O~\cite{Ren2015FasterRT}&Two-stage&14.9&69.05\\
			RoI Transformer~\cite{Ding2019LearningRT}&Two-stage&11.3&74.61\\
			S$^2$A-Net+MSCS~(Ours)&One-stage&14.2&\textbf{75.35}\\\hline			
		\end{tabular}
	}
	\label{tab_10}
\end{table}

\bibliographystyle{IEEEtran}
\bibliography{references.bib}

\vspace{11pt}

\vfill

\end{document}